\documentclass[11pt]{article}

\usepackage[preprint]{acl}

\usepackage{times}
\usepackage{latexsym}

\usepackage[T1]{fontenc}

\usepackage[utf8]{inputenc}

\usepackage{microtype}

\usepackage{inconsolata}

\usepackage{graphicx}
\usepackage{booktabs}
\usepackage{multirow}
\usepackage{multicol}
\usepackage{subcaption}
\usepackage{tabularx}
\usepackage[table]{xcolor}
\usepackage{amsmath}
\usepackage{tcolorbox}
\usepackage[utf8]{inputenc}
\usepackage[T1]{fontenc}
\usepackage{enumitem}
\usepackage{array}
\usepackage{float}
\newcolumntype{C}{>{\centering\arraybackslash}X}
\usepackage{polyglossia}
\setmainlanguage{english}
\setotherlanguage{bengali}

\newfontfamily\bengalifont[
    Path = ./ ,
    Extension = .ttf ,
    Script = Bengali
]{FreeSerif}
\title{MultiSoc-4D: A Benchmark for Diagnosing Instruction-Induced Label Collapse in Closed-Set LLM Annotation of Bengali Social Media}

\author{
 \textbf{Souvik Pramanik\textsuperscript{1,*}},
 \textbf{S.M. Riaz Rahman Antu\textsuperscript{1}},
 \textbf{Shak Mohammad Abyad\textsuperscript{1}},
 \\
 \textbf{Md. Ibrahim Khalil\textsuperscript{1}},
 \textbf{Md. Shahriar Hussain\textsuperscript{1}},
\\
\{\textit{souvik.pramanik, riaz.antu, shak.abyad, ibrahim.khalil03,}\\ \textit{ shahriar.hussain01}\}@northsouth.edu\\
 \textsuperscript{1}North South University, Dhaka, Bangladesh
\\
 \small{
   \textbf{\textsuperscript{*}Correspondence:} \href{mailto:souvik.pramanik@northsouth.edu}{souvik.pramanik@northsouth.edu}
 }
}

\begin{document}
\maketitle
\begin{abstract}
Annotation automation via Large Language Models (LLMs) is the core approach for scaling NLP datasets; however, LLM behavior with respect to closed-set instructions in low-resource languages has not been well studied. We present \textbf{MultiSoc-4D}, a Bengali social media dataset benchmark, which contains 58K+ social media comments from six sources annotated along four dimensions: category, sentiment, hate speech, and sarcasm. By employing a structured pipeline where ChatGPT, Gemini, Claude, and Grok individually annotate separate partitions, while sharing a common validation set of 20\%, we diagnose LLM behavior systematically. We discover a prevalent phenomenon called \textbf{``instruction-induced label collapse"}, wherein LLMs show a systematic preference towards fallback labels (Other, Neutral, No), leading to high agreement rates but under-detection of minority categories. For example, we find that LLMs failed to detect 79\% and 75\% of instances with hateful and sarcastic content compared to a human-calibrated reference. Furthermore, we prove that it represents a \textbf{``label agreement illusion"}, statistically validated via almost null Fleiss' Kappa ($\kappa \approx -0.001$) on sarcasm detection. Across 40+ LLMs, we benchmark this annotation bias propagation within the training pipeline, regardless of architectural differences. We release MultiSoc-4D as a diagnostic benchmark for annotation biases in Bengali NLP.
\end{abstract}


\section{Introduction}

\begin{figure}[!h]
    \centering
    \includegraphics[width=\linewidth]{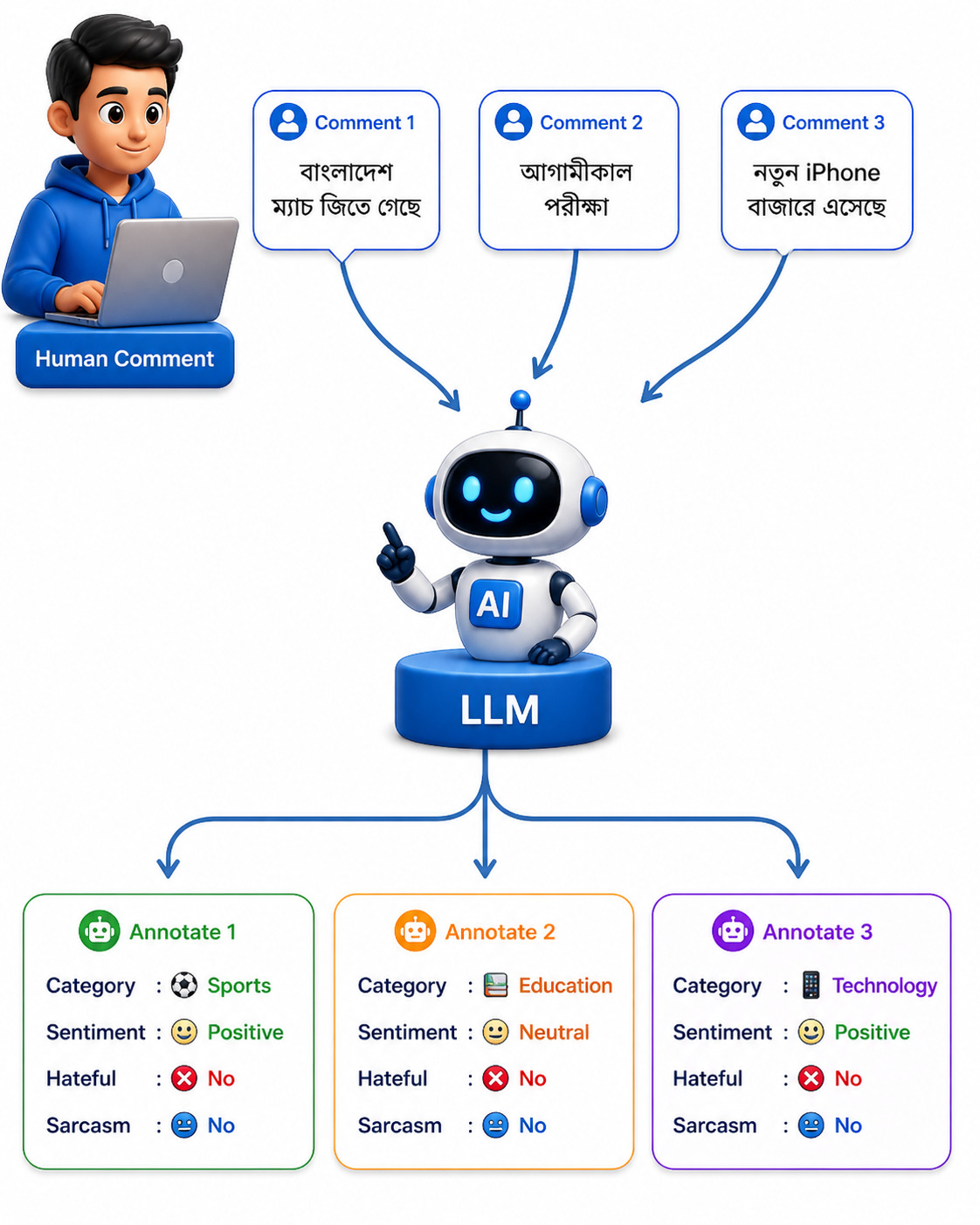}
    \caption{Exploring Bengali text annotation using LLMs.}
    \label{fig:intro-base}
\end{figure}

\begin{figure*}[htbp]
    \centering
    \includegraphics[width=\linewidth]{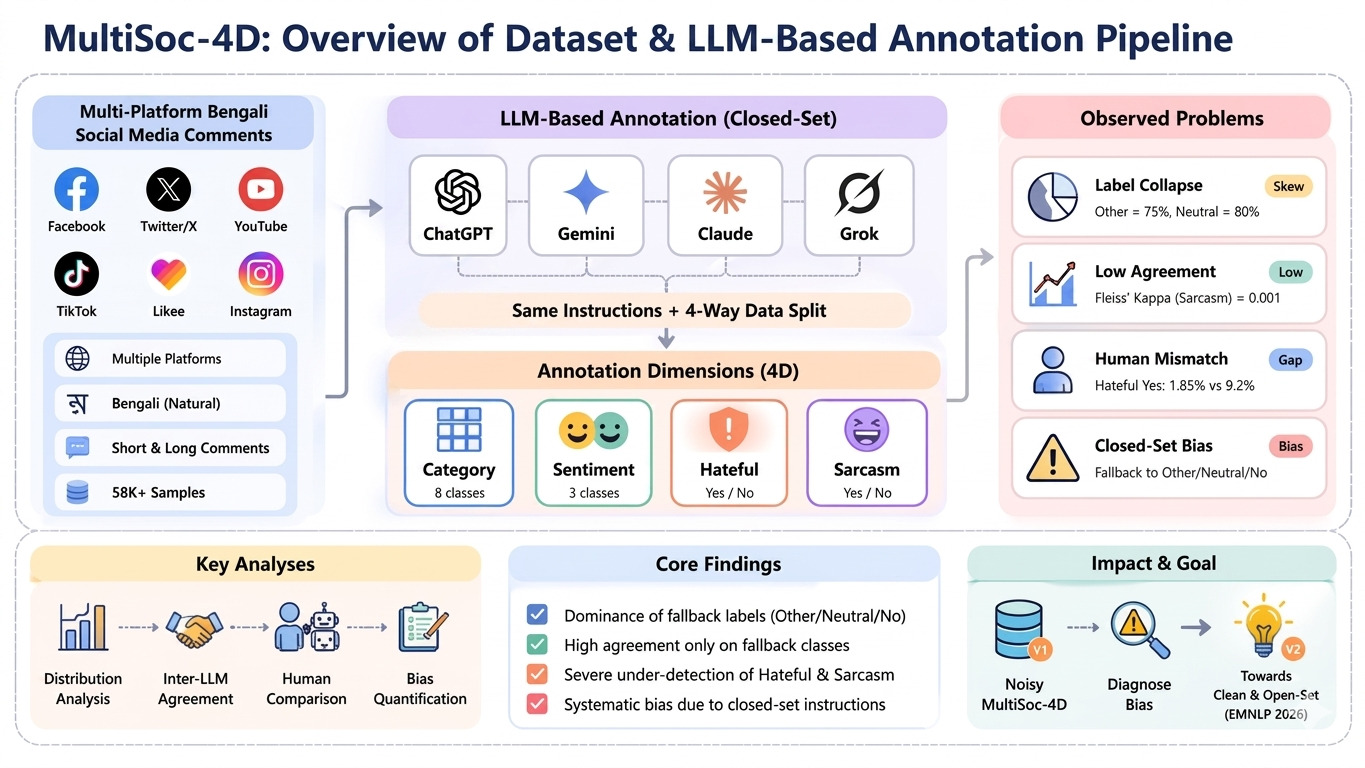}
    \caption{Overview of MultiSoc-4D dataset and LLM-based annotation pipeline.}
    \label{fig:intro}
\end{figure*}

The usage of Large Language Models (LLMs) as scalable substitutes for human annotators in building labeled datasets for a diverse set of NLP applications has been increasing. These models' good zero-shot and few-shot skills facilitate the creation of efficient and economical labeling pipelines \cite{brown2020language, ouyang2022training}. Consequently, LLM-driven annotation practices have emerged in various applications including sentiment analysis, hate-speech detection, and classification of topics. Despite the increase in the utilization of LLMs for these applications, not much is known about how the models operate as annotators. The existing literature concentrates on the performance of models without taking into consideration the role played by annotation guidelines and constrained label space in LLM decision-making processes \cite{wang2023self, gilardi2023chatgpt}.

Many annotation frameworks built on the basis of large language models assume a closed set labeling, implying that the prediction made for every observation belongs to a specific set of predefined labels. Many systems involve uncertainty-averse prompts, making the model choose from a set of fallback labels in case of low certainty (e.g., \textit{Other}, \textit{Neutral}, \textit{No}). Although such design was meant to make sure that the system remains consistent, this practice has a downside when applied to the labeling of data collected from social media platforms, which are characterized by diversity, ambiguity, and contextual dependency.

This study performed using a multilingual social media data in Bengali, \textbf{MultiSoc-4D}, annotated using multiple LLMs following a unified closed-set instruction approach. Our observations are described below:
\begin{itemize}
    \item \textbf{Label Collapse:} There is a tendency for LLM annotations to have a pronounced bias towards certain classes such as \textit{Others}, \textit{Neutral}, and \textit{No}.
    \item \textbf{Asymmetric Agreement:} While there is high agreement amongst LLMs for these particular labels, agreement in more sophisticated classes which do not appear often (such as sarcasm and hate speech) is almost non-existent.
    \item \textbf{Human-LLM Agreement Discrepancies:} As opposed to human annotations, there is evidence of a lack of understanding of implicit information by LLMs.
\end{itemize}
The above observations indicate that the level of agreement amongst the LLMs cannot be taken to mean semantic agreement and can rather be attributed to bias from instructions.

The key contributions of the study includs the following:
\begin{itemize}
    \item Presented \textbf{MultiSoc-4D}, a Bengali social media data set across multiple platforms labeled along four dimensions: category, sentiment, hatefulness, and sarcasm.
    \item Conducted an empirical investigation of LLM labeling using closed-set instructions, revealing issues of label collapse and skewing.
    \item Showed that agreement between LLMs is a deceptive measure of semantic consistency, as it mostly reflects the dominance of fallback labels.
    \item Estimated the gap between LLM and human annotations, demonstrating the challenges faced by LLMs in comprehending subtle cues.
\end{itemize}

The rest of the paper is structured as follows. In Section~\ref{sec:related}, we provide a literature review of prior work that uses LLMs for annotation and studies social media datasets. In Section~\ref{sec:dataset}, we introduce our dataset, MultiSoc-4D and present the LLM-based annotation system and its procedure. In Section~\ref{sec:analysis}, we perform an empirical analysis of LLM annotation behavior. In Section~\ref{sec:human}, we conduct human evaluation and quantify bias. The benchmarking and it's strategies under the biased annotated data is shown in Section~\ref{sec:bench}. Section~\ref{sec:dis} discusses the results and overall analysis. The limitation of the study is presented in Section~\ref{sec:lim} where we also mentioned our future direction. The ethical considerations are mentioned in Section ~\ref{sec:ec} and finally Section~\ref{sec:con} concludes the study.


\section{Related Work}
\label{sec:related}
\subsection{LLMs for Data Annotation}
\cite{hasan2024zerofewshotpromptingllms} present MUBASE as a new benchmark for sentiment analysis in Bengali language with 33,606 social media tweets and posts from Twitter and Facebook which are labeled as positive, negative, and neutral. The paper analyzes the results of fine-tuning of transformer models (such as BanglaBERT) and their comparison with zero- and few-shot prompting via GPT-4 and Flan-T5. The findings revealed that BanglaBERT outperforms the latter approach in terms of performance metrics; in particular, the former method achieved an F1-score of 69.39\% versus 61.17\% of the latter one. The research underlines the gaps of previously used datasets in terms of consistency in data annotations and lack of benchmarking for LLMs in the field. In turn, \cite{tan2024largelanguagemodelsdata} perform an overview of LLMs as tools for data annotation in independent and human-in-the-loop frameworks. The authors consider three ways of annotation – zero-/few-shot prompting, instruction-based labeling, and iterative improvement – and claim that the first two approaches allow reaching nearly human-level accuracy with reduced costs and time. However, LLMs exhibit sensitivity to prompting, are prone to biases and hallucinations especially in subjective and multilingual contexts, and have not been sufficiently evaluated in low-resource languages like Bangla yet.

\subsection{Social Media Data and Multi-Label Classification Datasets}
BANHATE \cite{raquib-etal-2025-banhate} is a dataset for Bangla hate speech classification, with 19,203 YouTube comments classified on a binary and fine-grained hate basis. This work assessed LLM and transformers and found LLaMA-3.1 (8B) with LoRA to give the best performance (83.83\% F1 for Hate). This dataset overcomes the shortcomings of the previous binary-only classification by allowing fine-grained and realistic hate detection. SentiGOLD \cite{sentigold_2023} presented an extensive Bangla dataset with seven thousand sentiments and a total of five different labels from different domains. It achieved a macro F1 score up to 0.62 with the use of deep and ML models. This dataset overcomes the drawbacks of noise in labeling and non-standard methods of previous data. 
BanglaBook \cite{kabir-etal-2023-banglabook} presented a very large dataset on Bangla books and their reviews of 158K. With Bangla-BERT achieving an F1 score of 93.31\%, this data set outperforms traditional models. Overcoming the problems of previous limited datasets, this allows for effective product sentiment classification. The work of \cite{paul2025analyzingemotionsbanglasocial} proposed EmoNoBa, a dataset of 22,698 annotated comments based on six emotions. The classical model performed better than BiLSTM (F1 = 38.69\%), while the best one was AdaBoost. LIME was used for explanation, solving the problem of interpretability in previous emotion classification datasets. \cite{10129187} introduced BE-CM, a Bangla-English code-mixed sentiment classification dataset consisting of 18,074 reviews. XGBoost with FastText and augmentation obtained an F1 score of 87\%, making it more robust to noisy texts and not requiring parallel corpora. \cite{das-bandyopadhyay-2010-labeling} created a sentence-level Bengali emotion recognition dataset by labeling sentences based on Ekman's six emotions, covering 12K sentences. The most successful method was SVM (accuracy = 80.55\%), allowing for nuanced analysis at the level of emotions. \cite{haider-etal-2025-banth} created BanTH, a dataset of 37,350 transliterated Bangla comments in seven classes of hate speech. The best model TB-mBERT reached 77.36\% Macro-F1, addressing transliteration and multi-label challenges in real-world data. Potrika (\cite{ahmad2022potrikarawbalancednewspaper}) was proposed as a 320K Bangla news dataset. The accuracy of GRU+FastText is 92\%, but the model's performance declined substantially with weak supervision, emphasizing the need for manual labeling. BanglishRev \cite{shamael2024banglishrevlargescalebanglaenglishcodemixed} consists of 1.74M e-commerce review datasets with Bangla, English, and Banglish text. BanglishBERT demonstrated F1 scores around 94\%, enabling large-scale sentiment and behavior analysis. It was proved in \cite{islam-etal-2022-emonoba} that classic models perform better than deep models in EmoNoBa (22K comments) because of the informal nature of the Bangla text compared to the pretrained language model. BnSentMix \cite{alam-etal-2025-bnsentmix} is a 20K code-mixed Bangla and English text dataset. Transformer-based models achieved 69.8\% accuracy, improving handling of real-world mixed-language content. \cite{article_hossain} created a small-scale emotion-based Bangla-English dataset (2,055 comments). SVM yielded an accuracy rate of 85.7\%, suggesting that emojis play a critical role in sentiment classification. \cite{HASAN2024111107} presented a new Bangla ASRB dataset; however, due to its small sample size and narrow scope, it cannot be used for benchmarking. Although this paper has contributed to fine-grained sentiment analysis, there is room for improvement. \cite{ISLAM2024100069} proposed a dataset named BangDSA containing a large number of 203K comments, along with 15 types of emotions. CNN-BiLSTM achieved accuracies of 90.24\% (15 classes) and 95.71\% (three classes); however, the data collection is unbalanced, and the dataset is not available. Hossain et al. \cite{hossain_fahima2025} developed a tiny Bangla sentiment dataset consisting of only 3K data with three algorithms: SVM, CNN, and LSTM. The LSTM model showed better performance (informal accuracy=80.3\%),

\subsection{Annotation Bias and Label Noise}
NC-SentNoB (\cite{elahi-etal-2024-comparative}) is a benchmark consisting of 15,000 Bangla samples labeled for 10 types of noise. \cite{elahi-etal-2024-comparative} conducted experiments using several models such as SVM, BiLSTM, Bangla-BERT, and MuRIL. Bangla-BERT-Base was found to outperform others in noise detection (F1: 0.62). The best sentiment performance was achieved by BanglaBERT (F1: 0.75), but it slightly dropped after noise reduction (F1: 0.73). In other words, current techniques for addressing label noise may not work well enough and change the semantic meanings. \cite{choi2024multinewscostefficientdatasetcleansing} presented Multi-news+, which is an LLM-based framework for cleaning datasets to save on labeling costs. Compared with heuristic methods, \cite{choi2024multinewscostefficientdatasetcleansing} showed that Multi-news+ led to higher annotation consistency and performance of machine learning models trained on cleaned datasets. Unfortunately, the framework's efficacy depends on the prompt design and language models used, being less effective in low-resource and domain-specific settings.

\subsection{Closed-set vs Open-set Annotation}
A cost-efficient LLM-based annotation approach for online dataset labeling proposed in \cite{elumar2025costawarellmbasedonlinedataset} involves balancing between LLM-based and less costly methods, thus achieving improved annotation efficiency without compromising performance. However, LLM-based bias is still present in the work, and ambiguity and domain specificity may negatively affect reliability, with insufficient evaluations performed on controlled datasets. \cite{bogdanov2024nunerentityrecognitionencoder} propose a novel framework called NuNER based on synthetic LLM-produced labels for pretraining encoders in the task of named entity recognition. Token-level performance metrics exceed zero-shot and supervised models, yet the proposed framework requires high-quality annotations, is sensitive to labeling bias, and performs only on English datasets. \cite{electronics14142800} introduce a unified LLM-based annotation platform, utilizing Llama 3.3 with different attribution techniques, including zero-shot, few-shot, chain-of-thought, and role-based prompts for generating synthetic datasets and labels. Although performance is impressive (99\% on synthetic data and up to 92\% on AG News), scalability and context window problems, along with low effectiveness for similar classes, exist.

The summary of related work on Bangla NLP Datasets and LLM-based Annotation is shown in Table~\ref{tab:related_work}.

\begin{table*}[htbp]
\centering
\small
\renewcommand{\arraystretch}{1.2}
\begin{tabular}{p{3.2cm} p{3.2cm} p{3.2cm} p{5.2cm}}
\hline
\cellcolor{gray!15}\textbf{Work} & \cellcolor{gray!15}\textbf{Task / Domain} & \cellcolor{gray!15}\textbf{Dataset Size / Type} & \cellcolor{gray!15}\textbf{Key Contribution \& Limitation} \\
\hline

MUBASE \cite{hasan2024zerofewshotpromptingllms} 
& Sentiment Analysis (Bangla) 
& 33K social media posts 
& Benchmarks LLM prompting vs fine-tuned models; BanglaBERT outperforms GPT-4. Limitation: LLM underperformance in low-resource setup. \\

SentiGOLD \cite{sentigold_2023} 
& Sentiment Analysis (Multi-domain) 
& 70K samples, 5-class 
& Gold-standard Bangla sentiment dataset; improves annotation quality. Limitation: moderate performance ceiling. \\

EmoNoBa \cite{islam-etal-2022-emonoba} 
& Emotion Classification 
& 22K social media comments 
& Shows classical models outperform deep models due to domain mismatch. Limitation: informal text challenges. \\

BE-CM \cite{10129187} 
& Code-mixed Sentiment 
& 18K reviews 
& Handles Bangla-English mixing using augmentation; XGBoost best. Limitation: limited dataset size. \\

BanTH \cite{haider-etal-2025-banth} 
& Hate Speech (Transliterated Bangla) 
& 37K YouTube comments 
& First large transliterated multi-label hate dataset. Limitation: domain restricted to YouTube. \\

BANHATE \cite{raquib-etal-2025-banhate} 
& Hate Speech (Fine-grained) 
& 19K comments 
& Fine-grained hate categories + LLM evaluation. Limitation: LLM bias persists. \\

BanglishRev \cite{shamael2024banglishrevlargescalebanglaenglishcodemixed} 
& E-commerce Sentiment 
& 1.74M reviews 
& Large-scale Bangla-English mixed dataset; BanglishBERT strong performance. Limitation: weak supervision noise. \\

NC-SentNoB \cite{elahi-etal-2024-comparative} 
& Noisy Text Analysis 
& 15K samples, 10 noise types 
& Studies noise impact on sentiment models. Limitation: noise reduction may distort meaning. \\

Multi-news+ \cite{choi2024multinewscostefficientdatasetcleansing} 
& Dataset Cleansing 
& Multi-domain corpora 
& LLM-based automated data cleaning. Limitation: prompt bias and low-resource weakness. \\

NuNER \cite{bogdanov2024nunerentityrecognitionencoder} 
& Named Entity Recognition 
& Synthetic LLM-labeled data 
& Improves NER via LLM-generated annotations. Limitation: English-only evaluation. \\

Llama Annotation Framework \cite{electronics14142800} 
& Dataset Generation / Annotation 
& AG News + synthetic corpora 
& Unified LLM-based labeling system (up to 99\% F1 synthetic). Limitation: scalability + ambiguity issues. \\

BangDSA \cite{ISLAM2024100069} 
& Emotion Classification 
& 203K social media comments 
& Large 15-class emotion dataset with strong performance. Limitation: not publicly available + imbalance. \\

BnSentMix \cite{alam-etal-2025-bnsentmix} 
& Code-mixed Sentiment 
& 20K posts 
& Handles Bangla-English mixed social media text. Limitation: moderate performance ceiling. \\

\rowcolor{blue!8}
\textbf{MultiSoc-4D (Ours)}
& Multi-task: Category, Sentiment, Hate Speech, Sarcasm (Bangla)
& 58K+ multi-platform comments (Facebook, Twitter/X, YouTube, TikTok, Likee, Instagram)
& First Bengali social media benchmark diagnosing instruction-induced label collapse in closed-set LLM annotation. Introduces Bias Ratio and agreement illusion analysis across 4 annotator LLMs and 40+ downstream models. Limitation: silver-standard labels; human calibration is future work.\\
\hline
\end{tabular}
\caption{Summary of Related Work on Bangla NLP Datasets and LLM-based Annotation}
\label{tab:related_work}
\end{table*}


\section{MultiSoc-4D Dataset}
\label{sec:dataset}
\subsection{Data Collection}
\subsubsection{Data Sources}
The MultiSoc-4D dataset was built using posts from different social media websites such as Facebook, Twitter (X), YouTube, TikTok, Likee, and Instagram. The selection of social media platforms was done to maintain heterogeneity in terms of writing style, user base, and discourse. In contrast to specialized domain corpora, social media posts contain informal speech, slang, code switching, and context-based meanings. The posts of the dataset are predominantly written in Bengali, although there might be instances of English words, hashtags, and user tags.

\begin{figure*}[t]
    \centering
    \includegraphics[width=\linewidth]{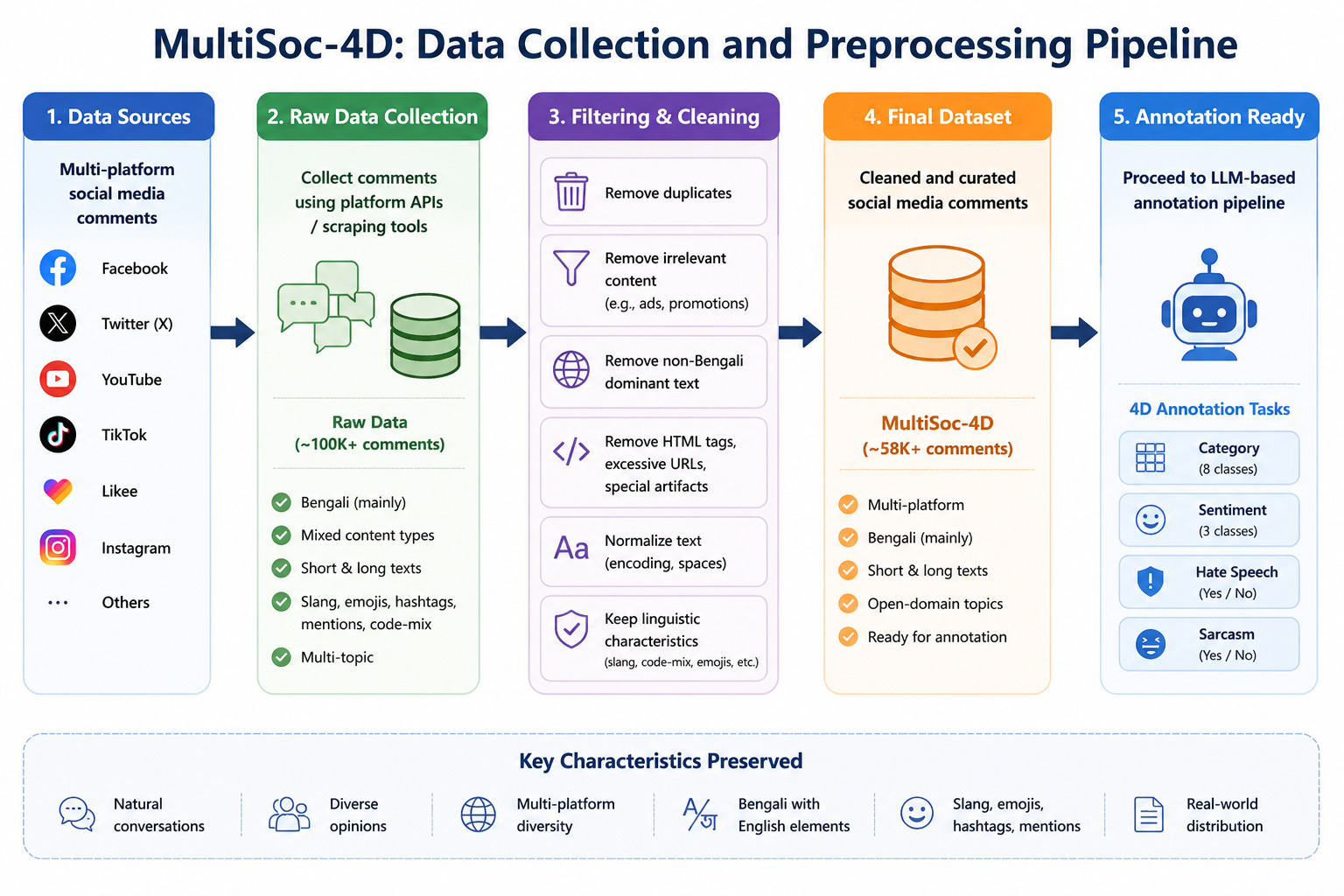}
    \caption{Data collection and preprocessing pipeline for MultiSoc-4D.}
\end{figure*}

\subsubsection{Sampling Strategy}
A random sampling method was employed in order to reflect the natural distribution without imposing balance among domains. The aim here is to maintain the natural attributes of social media datasets, including imbalanced classes, diverse topics, and varied languages. In order to prevent any bias due to oversampling from one particular source, proportional sampling was conducted across all sources. Content filtering based on categories was not performed manually.

\subsection{Dataset Schema}
\begin{figure}[t]
    \centering
    \includegraphics[width=\linewidth]{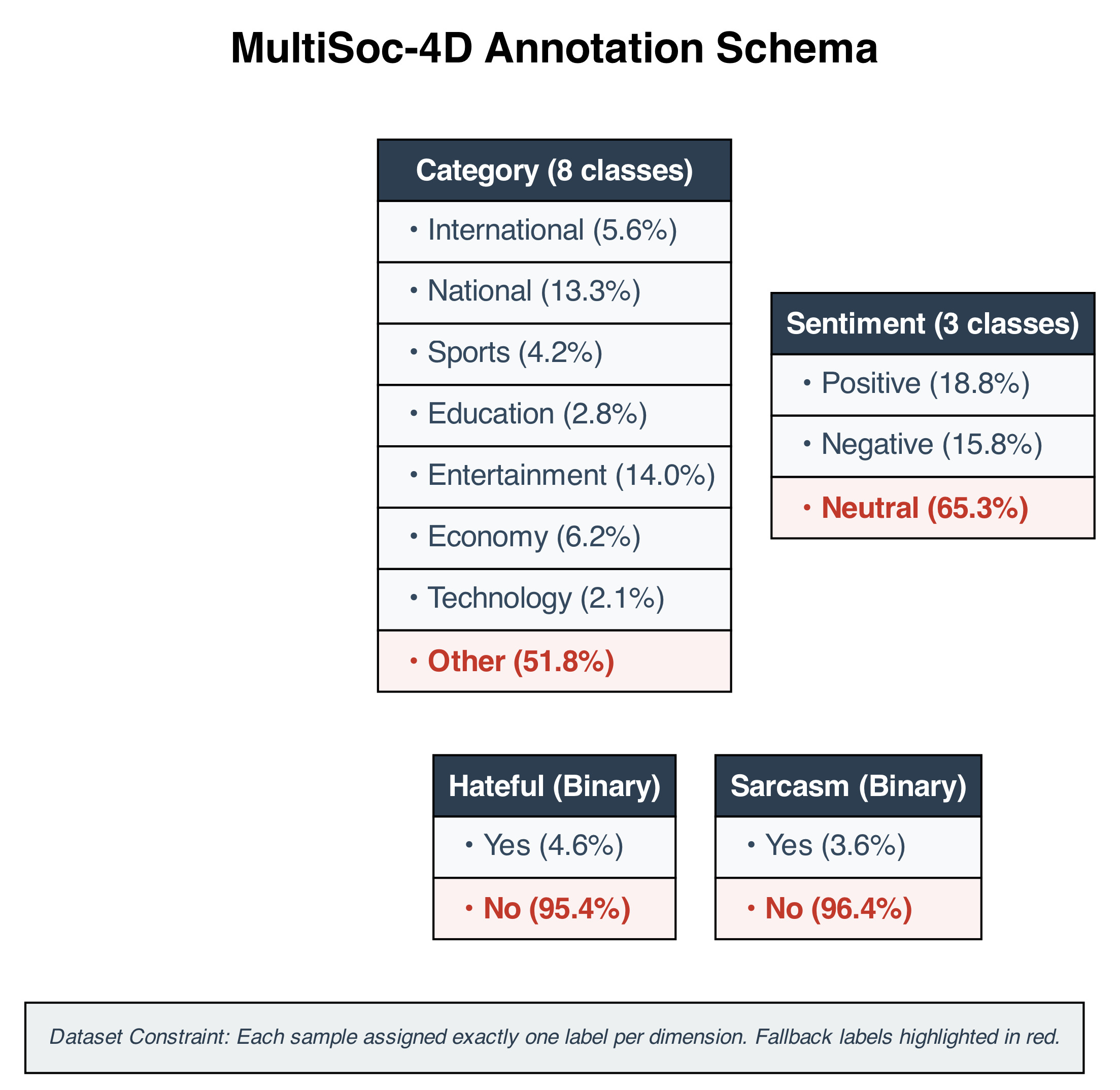}
    \caption{Dataset schema of MultiSoc-4D across four dimensions.}
    \label{fig:schema}
\end{figure}

The dataset is annotated across four dimensions: \textit{category}, \textit{sentiment}, \textit{hateful}, and \textit{sarcasm}. Figure~\ref{fig:schema} provides an overview of the dataset schema.

\subsubsection{Categories (8 Classes)}
Each sample is labeled by one of the eight exclusive categories:
\textit{International, National, Sports, Education, Entertainment, Economy, Technology,} and \textit{Others}. The \textit{Others} category is used as a catch-all category in case any of the samples do not fit into defined topics.

\subsubsection{Sentiment (3 Classes)}
The sentiment is labeled with one of the following labels:
\textit{Positive, Negative,} and \textit{Neutral}. The label \textit{Neutral} is assigned to the objectively neutral content as well as sentiment-neutral samples.

\subsubsection{Hate Speech Labeling}
This feature is labeled in a binary classification task:
\textit{Yes} and \textit{No}. The former is applied to samples containing direct or indirect hate speech directed to individuals or groups.

\subsubsection{Sarcasm Labeling}
Similarly to Hate Speech, the Sarcasm feature is also labeled with Yes/No:
\textit{Yes} and \textit{No}. It characterizes implicit expressions, which diverge in literal meaning.

\subsection{Annotation Framework}

\subsubsection{Annotators}

The annotation process uses four large language models, namely ChatGPT\cite{gilardi2023chatgpt}, Gemini\cite{team2023gemini}, Claude\cite{anthropic2024claude}, and Grok\cite{xai2024grok}. These language models are chosen because of their high proficiency in instruction-following and text classification tasks.

\subsubsection{Instructions Design}

The annotators are provided with identical instructions for annotation purposes. In line with the closed-set labeling procedure, each dimension is assigned one label within a predefined set.

\textbf{Closed-Set Rules:}
For dealing with uncertainties in the annotation process, some default rules are included in the instructions:
\begin{itemize}
    \item Assign \textit{Other} for unclear categories
    \item Assign \textit{Neutral} for unclear sentiments
    \item Assign \textit{No} for unknown cases of hatefulness and sarcasm
\end{itemize}

Though this set of instructions is designed to facilitate consistency in the annotation process, it inadvertently introduces bias towards conservative labeling. The Extended Instructions are shown in \ref{sec : AG}.

\subsection{Dataset Statistics}

\subsubsection{Platform Distribution}
The dataset includes contributions from multiple social media platforms, with varying proportions. This diversity ensures coverage of different discourse styles and interaction patterns.
\begin{figure}[t]
    \centering
    \includegraphics[width=\linewidth]{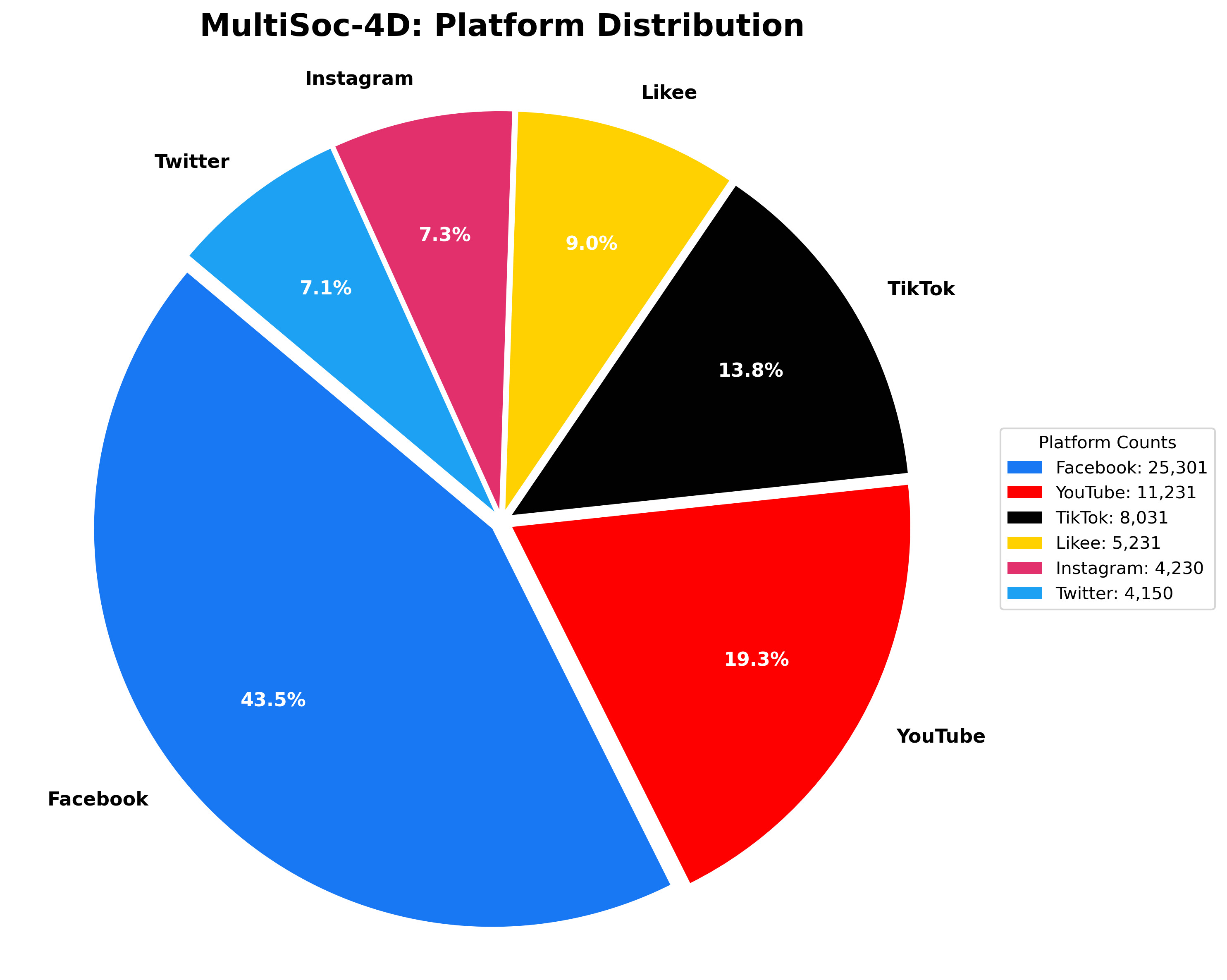}
    \caption{Class distribution across all the platforms.}
    \label{fig:distribution}
\end{figure}

Extra statistics are shown in \ref{sec: DE}.

\subsection{Data Preprocessing}

Before starting with the annotation phase, very few preprocessing activities are performed on the collected dataset so that its inherent nature is not lost. These activities are as follows:

\begin{itemize}
    \item Elimination of duplicates
    \item Simple normalization of text encoding
    \item Elimination of non-text elements like HTML, etc.
\end{itemize}

One key point is that the natural linguistic phenomena of the text, including the use of slang, colloquial language, and even minor code mixing, have been preserved to keep the essence of the data alive.

\section{Empirical Analysis of Annotation Behavior}
\label{sec:analysis}
\begin{figure*}[t]
    \centering
    \includegraphics[width=\linewidth]{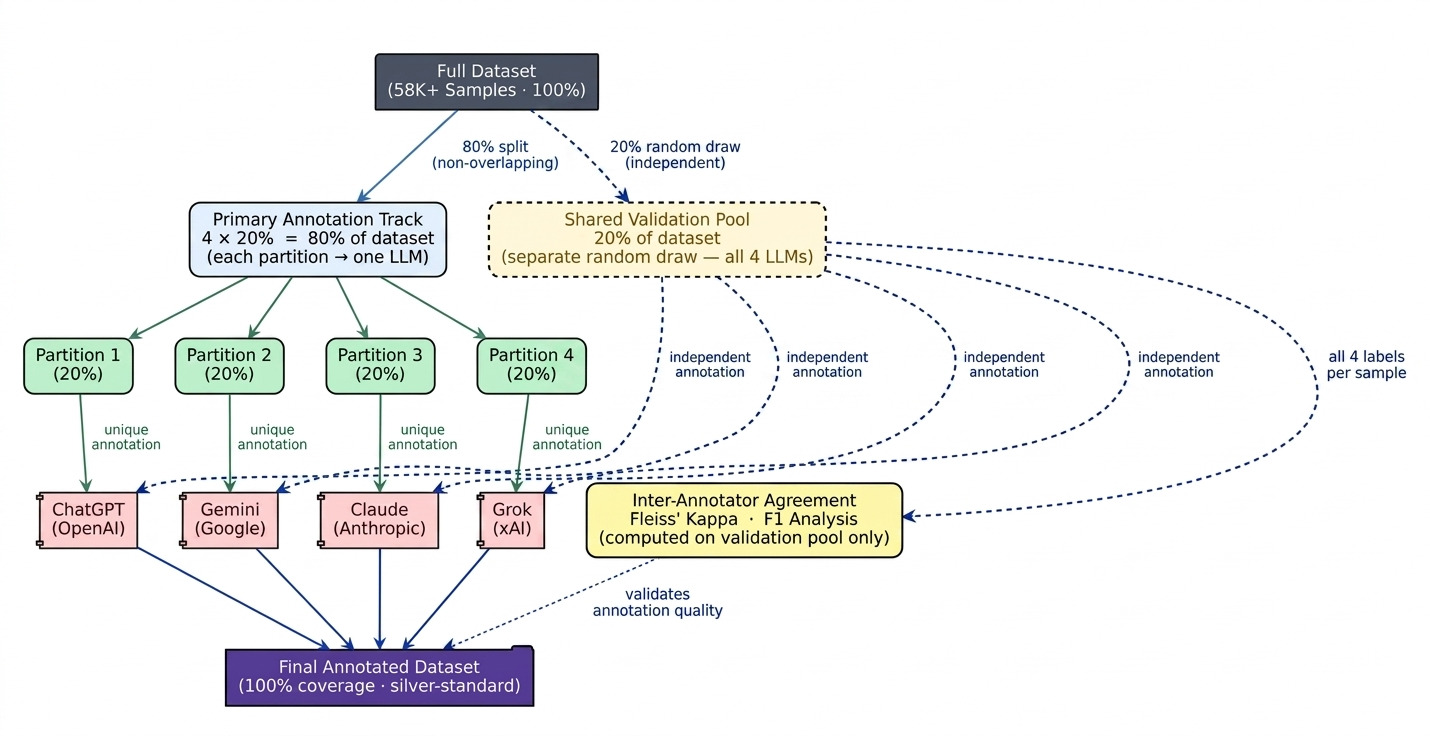}
    \caption{LLM-based annotation pipeline with dataset splitting and multi-model labeling.}
    \label{fig:pipeline}
\end{figure*}

The complete data set comprising N samples is split into four mutually exclusive subsets, each of which contains an equal number of samples. One of the subsets is then annotated by one of the LLMs. In order to conduct inter-annotator agreement studies, a 20\% representative sample from the complete data set — selected from all four subsets — is annotated by all four models independently. This results in a common validation set, where all four LLMs annotate the same samples."
Figure~\ref{fig:pipeline} illustrates the annotation pipeline.

\subsection{Label Distribution Collapse}

\begin{figure}[t]
    \centering
    \includegraphics[width=\linewidth]{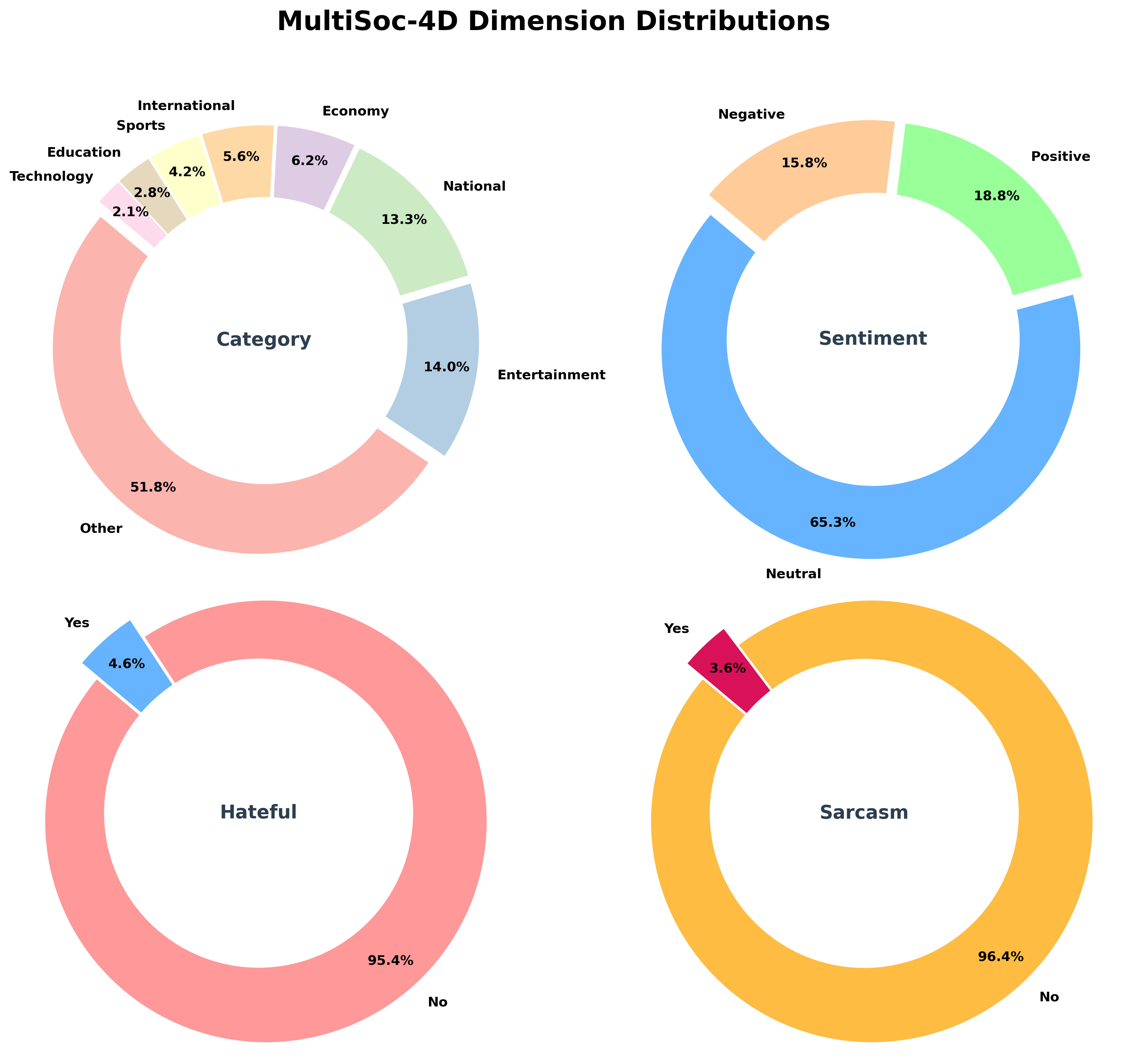}
    \caption{Label distribution across annotation dimensions showing strong fallback dominance.}
    \label{fig:label_dist}
\end{figure}

We start with the analysis of label distributions yielded by all LLM annotators in terms of four annotation dimensions: category, sentiment, hateful content, and sarcasm. According to Figure~\ref{fig:label_dist}, all models have strong and stable skew towards fallback labels. To be precise, the \textit{Other} label accounts for roughly 51.8\% of category labels, and \textit{Neutral} is predominant in sentiment annotations (65.3\%). Binary dimensions have an even stronger bias, with \textit{No} chosen for 95.4\% of hateful content samples and 96.4\% of sarcastic samples. That means that there is a serious compression of the label space, during which ambiguous input data gets mapped into conservative fallback classes. Moreover, this tendency is universal for all models considered, which suggests a lack of specificity of the collapse process and its dependence on instruction formulation rather than individual model properties.

\subsection{Cross-Model Consistency}

The following analysis will focus on identifying any consistency between various large language models when presented with the same labeling task under identical conditions. Even though there is variability in both the architecture used for the LLM and its training dataset, ChatGPT\cite{gilardi2023chatgpt}, Gemini\cite{team2023gemini}, Claude\cite{anthropic2024claude}, and Grok\cite{xai2024grok} all have an extremely similar label distribution on all dimensions. The underlying structure for all four LLMs appears to be quite similar, characterized by a very high dependence on using fallback labels while avoiding minority semantic labels.

\begin{figure}[t]
    \centering
    \includegraphics[width=\linewidth]{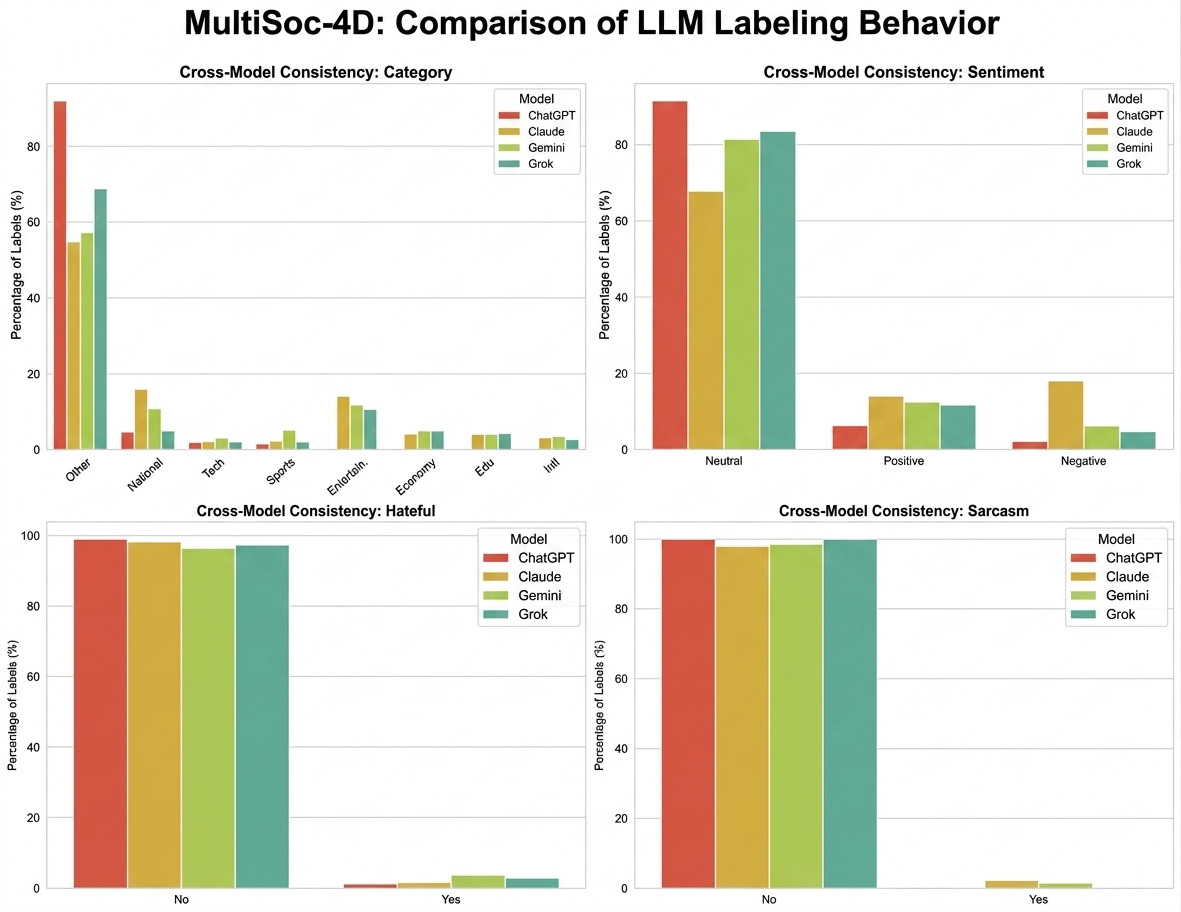}
    \caption{All models reliance on fallback labels.}
    \label{fig:rel_dist}
\end{figure}

This consistency shown in Figure~\ref{fig:rel_dist} indicates that closed-set annotation induces a shared behavioral regime across heterogeneous LLMs, effectively overriding model-level variability.


\subsection{Inter-Annotator Agreement Analysis}

We quantify inter-model agreement using Fleiss’ Kappa over a randomly sampled 20\% subset of the dataset, where all four models annotate the same instances.

\subsubsection{Overall Fleiss' Kappa}

It is evident that there is great variation in the level of agreement between annotators based on annotation dimension, as illustrated in Table \ref{tab:fleiss_kappa_results}. For instance, we have moderate agreement when it comes to category ($\kappa \approx 0.41$) and sentiment ($\kappa \approx 0.55$), but the level of agreement is rather poor in the case of hateful content ($\kappa < 0.39$). There is almost no agreement at all in terms of sarcasm ($\kappa \approx -0.001$).

\begin{table}[h]
    \centering
    \caption{Fleiss' Kappa Scores Across Annotation Dimensions}
    \label{tab:fleiss_kappa_results}
    \begin{tabular}{@{}lc p{2.5cm}@{}}
        \toprule
        \cellcolor{gray!15}\textbf{Dimension} & \cellcolor{gray!15}\textbf{Fleiss' $\kappa$} & \cellcolor{gray!15}\textbf{Agreement Strength} \\ \midrule
        Sentiment          & 0.5553                      & Moderate                   \\
        Category           & 0.4091                      & Moderate                   \\
        Hateful Content    & < 0.3868                    & Slight / Poor              \\
        Sarcasm            & -0.0011                  & Negligible                 \\ \bottomrule
    \end{tabular}
\end{table}

This discrepancy shown in Table~\ref{tab:fleiss_kappa_results} suggests that LLMs exhibit apparent consistency only in coarse-grained or structurally simple tasks, while failing to maintain alignment in semantically complex or context-dependent dimensions.

\subsubsection{Class-wise Agreement}

\begin{table}[h]
    \centering
    \caption{Class-wise Agreement Metrics Across Four Annotators. (FAR = Full Agreement Ration, AAS = Average Agreement Strength.)}
    \label{tab:class_agreement_unified}
    \addtolength{\tabcolsep}{-3pt} 
    \small
    \begin{tabular}{llcc}
        \toprule
        \cellcolor{gray!15}\textbf{Dimension} & \cellcolor{gray!15}\textbf{Class} & \cellcolor{gray!15}\textbf{FAR} & \cellcolor{gray!15}\textbf{AAS} \\ \midrule
        \multirow{4}{*}{\shortstack[l]{Category\\($\kappa=0.41$)}} 
        & \cellcolor{blue!15}Other      & \cellcolor{blue!15}0.4531 & \cellcolor{blue!15}0.7270 \\
        & Technology       & 0.2975 & 0.5881 \\
        & Sports     & 0.2095 & 0.4556 \\
        & Economy    & 0.0000 & 0.5715 \\
        & Education  &0.0000   &0.5195 \\
        &Entertainment  &0.0000 &0.5307\\
        &International  &0.0000     &0.4080 \\ 
        &National &0.0861  &0.4116\\\midrule
        \multirow{3}{*}{\shortstack[l]{Sentiment\\($\kappa=0.56$)}} 
        & \cellcolor{blue!15}Neutral    & \cellcolor{blue!15}0.7008 & \cellcolor{blue!15}0.8763 \\
        & Positive   & 0.3359 & 0.7018 \\
        & Negative   & 0.0992 & 0.3891 \\ \midrule
        \multirow{2}{*}{\shortstack[l]{Hateful\\($\kappa=0.39$)}} 
        & \cellcolor{blue!15}No         & \cellcolor{blue!15}0.9498 & \cellcolor{blue!15}0.9776 \\
        & Yes        & 0.0135 & 0.4535 \\ \midrule
        \multirow{2}{*}{\shortstack[l]{Sarcasm\\($\kappa=-0.001$)}} 
        & \cellcolor{blue!15}No         & \cellcolor{blue!15}0.9612 & \cellcolor{blue!15}0.9902 \\
        & Yes        & 0.0000 & 0.2533 \\ \bottomrule
    \end{tabular}
\end{table}

The findings from Table~\ref{tab:class_agreement_unified} reveal a clear bias towards negative or neutral default predictions. On the Sarcasm and Hatefulness dimensions, the models tend to reach agreement on whether or not the feature is present (choosing "No") almost consistently; however findings suggest that while the annotators (LLM models) effectively filter "standard" content, the detection of specific topics and sentiments remains highly subjective and prone to disagreement.


\subsection{Agreement vs Label Frequency}

We further analyze the relationship between label frequency and agreement. A strong positive correlation is observed between class prevalence and inter-annotator agreement.

\begin{figure}[t]
    \centering
    \includegraphics[width=\linewidth]{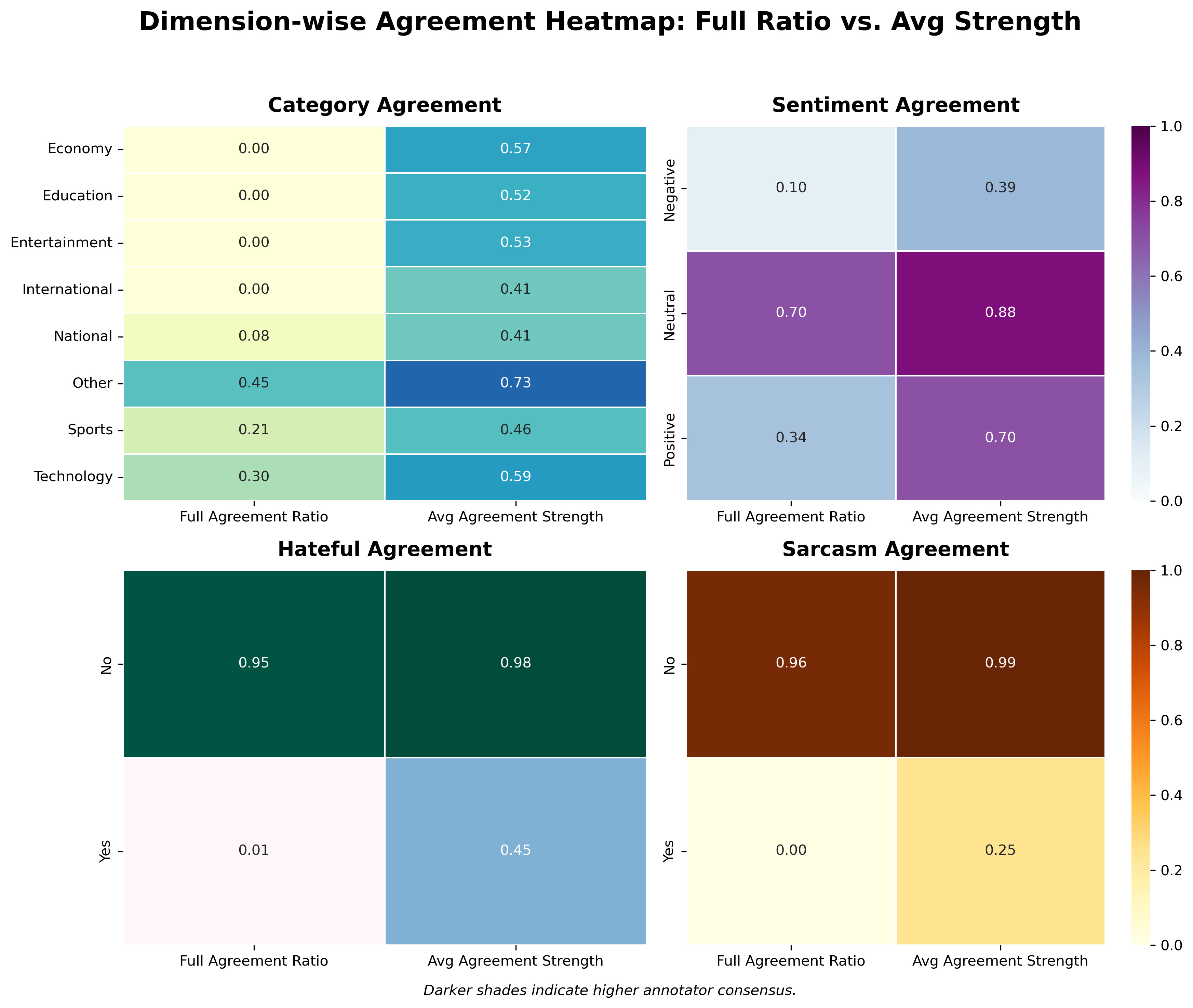}
    \caption{Inter-annotator agreement (Fleiss' Kappa) across Label Frequency.}
    \label{fig : kap}
\end{figure}

From Figure~\ref{fig : kap} it is clear that high-frequency fallback labels consistently exhibit higher agreement, while low-frequency semantic classes show unstable and near-random agreement behavior. This suggests that agreement metrics in closed-set LLM annotation are heavily confounded by label distribution imbalance rather than reflecting true inter-model understanding.

\section{Human Evaluation and Bias Quantification}
\label{sec:human}
This section compares LLM-generated annotations with human annotations on a stratified random subset of 500 samples (Which is called the GOLD set). The goal is to quantify distributional bias and identify systematic deviations in LLM labeling behavior.

\subsection{Human Annotation Setup}

A subset of 500 samples is independently annotated by human annotators following the same four-dimensional schema used for LLM annotation. Each instance is labeled for category, sentiment, hateful content, and sarcasm according to the original annotation guidelines. The instructions for the human and the LLMs were same for the annotation criteria. The instructions are clearly mentioned in \ref{sec : AG}.

This subset serves as a calibration set to evaluate distributional and error-level divergence between human judgments and LLM-generated annotations. Actually this is working as a validation of the problem raised in this study.


\subsection{Distribution Comparison}

We compare the label distributions produced by LLMs (averaged across models) and human annotations on the 500-sample subset.

\begin{table}[h!]
\centering
\caption{Comparative Label Distribution: Individual LLMs vs. Human Gold Standard (N=500). Cl = ClaudeAI, GPT = ChatGPT, Ge = Gemini, Gr = Grok, H = Human.}
\label{tab:distribution_full_comparison}
\small
\addtolength{\tabcolsep}{-4pt} 
\begin{tabular}{@{}llccccc@{}}
\toprule
\cellcolor{gray!15}\textbf{Dim.} & \cellcolor{gray!15}\textbf{Class} & \cellcolor{gray!15}\textbf{Cl.} & \cellcolor{gray!15}\textbf{GPT} & \cellcolor{gray!15}\textbf{Ge.} & \cellcolor{gray!15}\textbf{Gr.} & \cellcolor{cyan!10}\textbf{H.} \\ \midrule
\multirow{8}{*}{Category} & \cellcolor{blue!10}Other & \cellcolor{blue!10}169 & \cellcolor{blue!10}427 & \cellcolor{blue!10}390 & \cellcolor{blue!10}441 & \cellcolor{blue!10}291 \\
 & National & 191 & 47 & 51 & 27 & \cellcolor{cyan!10}96 \\
 & Entertainment & 54 & 7 & 12 & 11 & \cellcolor{cyan!10}29 \\
 & International & 23 & 8 & 7 & 5 & \cellcolor{cyan!10}24 \\
 & Technology & 25 & 2 & 4 & 2 & \cellcolor{cyan!10}20 \\
 & Economy & 19 & 4 & 11 & 6 & \cellcolor{cyan!10}18 \\
 & Sports & 13 & 3 & 10 & 3 & \cellcolor{cyan!10}12 \\
 & Education & 6 & 2 & 15 & 5 & \cellcolor{cyan!10}10 \\ \midrule
\multirow{3}{*}{Sentiment} & \cellcolor{blue!10}Neutral & \cellcolor{blue!10}173 & \cellcolor{blue!10}446 & \cellcolor{blue!10}460 & \cellcolor{blue!10}470 & \cellcolor{blue!10}316 \\
 & Negative & 231 & 23 & 29 & 19 & \cellcolor{cyan!10}90 \\
 & Positive & 96 & 31 & 11 & 11 & \cellcolor{cyan!10}94 \\ \midrule
\multirow{2}{*}{Hateful} & \cellcolor{blue!10}No & \cellcolor{blue!10}474 & \cellcolor{blue!10}491 & \cellcolor{blue!10}489 & \cellcolor{blue!10}490 & \cellcolor{blue!10}454 \\
 & Yes & 26 & 9 & 11 & 10 & \cellcolor{cyan!10}46 \\ \midrule
\multirow{2}{*}{Sarcasm} & \cellcolor{blue!10}No & \cellcolor{blue!10}467 & \cellcolor{blue!10}488 & \cellcolor{blue!10}498 & \cellcolor{blue!10}493 & \cellcolor{blue!10}465 \\
 & Yes & 33 & 12 & 2 & 7 & \cellcolor{cyan!10}35 \\ \bottomrule
\end{tabular}
\end{table}

The results from Table~\ref{tab:distribution_full_comparison} indicate that LLMs are currently "risk-averse" classifiers. Their tendency to favor Neutral, Non-Hateful, and Non-Sarcastic labels leads to a sanitized version of the data that lacks the granularity and sensitivity of human judgment. This poses a significant challenge for using LLMs in automated content moderation and nuanced social media analysis.


\subsection{Bias Ratio Analysis}

In order to measure how far apart the outputs from the model are compared to the ground truth human labels, we introduce the Bias Ratio metric, defined by
\begin{equation}
    \text{Bias Ratio} = \frac{\text{LLM Label Frequency (Avg.)}}{\text{Human Label Frequency}}.
\end{equation}

As illustrated in Table \ref{tab:bias_ratio_analysis}, the analysis of the Bias Ratio demonstrates a pattern of inherent structural bias within the LLM labels. The following table shows that fallback classes are significantly overrepresented ($\text{Bias Ratio}>1$) across all dimensions while semantic and minority classes such as Sarcasm, Hateful Content, and Sports, on the other hand, are persistently underrepresented ($\text{Bias Ratio}<1$).

\begin{table}[h]
    \centering
    \caption{Bias Ratio Across All Annotation Dimensions ($N=500$). H = Human, BR = Bias Ratio.}
    \label{tab:bias_ratio_analysis}
    \addtolength{\tabcolsep}{-3pt}
    \small
    \begin{tabular}{llccc}
        \toprule
        \cellcolor{gray!15}\textbf{Dim.} & \cellcolor{gray!15}\textbf{Class} & \cellcolor{gray!15}\textbf{LLM Avg.} & \cellcolor{gray!15}\textbf{H.} & \cellcolor{gray!15}\textbf{BR.} \\ \midrule
        \multirow{8}{*}{Category} & \cellcolor{red!25}Other         & 356.75 & 291 & \cellcolor{red!25}1.22 \\
                                   & National      & 79.00  & 96  & 0.82 \\
                                   & Entertainment & 21.00  & 29  & 0.72 \\
                                   & International & 10.75   & 24  & 0.45 \\
                                   & Technology    & 8.25   & 20  & 0.41 \\
                                   & Economy       & 10.00  & 18  & 0.56 \\
                                   & Sports        & 7.25   & 12  & 0.60 \\
                                   & Education     & 7.00   & 10  & 0.70 \\ \midrule
        \multirow{3}{*}{Sentiment} & \cellcolor{red!25}Neutral       & 387.25 & 316 & \cellcolor{red!25}1.23 \\
                                   & Negative      & 75.50  & 90  & 0.84 \\
                                   & Positive      & 37.25  & 94  & 0.40 \\ \midrule
        \multirow{2}{*}{Hateful}   & Yes           & 14   & 46  & 0.30 \\
                                   & \cellcolor{red!25}No            & 486 & 454 & \cellcolor{red!25}1.07 \\ \midrule
        \multirow{2}{*}{Sarcasm}   & Yes           & 13.5   & 35  & 0.38 \\
                                   & \cellcolor{red!25}No            & 486.5 & 465 & \cellcolor{red!25}1.04 \\ \bottomrule
    \end{tabular}
\end{table}

Bias Ratios provide an estimate of how much "filtering" LLMs do. For sarcasm (yes) and hateful (yes), bias ratios are 0.25 and 0.20 respectively, which shows that LLMs pick up merely a quarter or fifth of what is detected by human beings. On the other hand, Other and neutral have bias ratios of 1.29 and 1.27, respectively, which show that they push away labels from more informative classes.


\subsection{Model to Human Distribution}
In order to calculate the Model to Human Distribution we applied the formula : 
\begin{equation}
    \text{M2H Distribution} = \frac{\text{LLM Label Count}}{\text{Human Label Count}}.
\end{equation}

\begin{table}[h!]
\centering
\caption{Model to Human(M2H) Distributional Ratios (Values closer to 1.00 are better). Cl = ClaudeAI, GPT = ChatGPT, Ge = Gemini, Gr = Grok, H = Human.}
\label{tab:model_human_ratio}
\small
\addtolength{\tabcolsep}{-4pt}
\begin{tabular}{@{}llcccc@{}}
\toprule
\cellcolor{gray!15}\textbf{Dim.} & \cellcolor{gray!15}\textbf{Class} & \cellcolor{green!25}\textbf{Cl.} & \cellcolor{gray!15}\textbf{GPT} & \cellcolor{gray!15}\textbf{Ge.} & \cellcolor{gray!15}\textbf{Gr.} \\ \midrule
\multirow{8}{*}{Category} & Other & 0.58 & 1.47 & 1.34 & 1.52 \\
 & National & \cellcolor{red!25}1.99 & 0.49 & 0.53 & 0.28 \\
 & Entertainment & 1.86 & 0.24 & 0.41 & 0.38 \\
 & International & \cellcolor{green!25}0.96 & 0.33 & 0.29 & 0.21 \\
 & Technology & \cellcolor{green!15}1.25 & 0.10 & 0.20 & 0.10 \\
 & Economy & \cellcolor{green!25}1.06 & 0.22 & 0.61 & 0.33 \\
 & Sports & \cellcolor{green!25}1.08 & 0.25 & 0.83 & 0.25 \\
 & Education & 0.60 & 0.20 & 1.50 & 0.50 \\ \midrule
\multirow{3}{*}{Sentiment} & Neutral & 0.55 & 1.41 & 1.46 & 1.49 \\
 & Negative & \cellcolor{red!25}2.57 & 0.26 & 0.32 & 0.21 \\
 & Positive & \cellcolor{green!25}1.02 & 0.33 & 0.12 & 0.12 \\ \midrule
\multirow{2}{*}{Hateful} & Yes & 0.57 & 0.20 & 0.24 & 0.22 \\
 & No & \cellcolor{green!25}1.04 & 1.08 & 1.08 & 1.08 \\ \midrule
\multirow{2}{*}{Sarcasm} & Yes & \cellcolor{green!25}0.94 & 0.34 & 0.06 & 0.20 \\
 & No & \cellcolor{green!25}1.00 & 1.05 & 1.07 & 1.06 \\ \bottomrule
\end{tabular}
\end{table}

Based on the distributional ratios and the raw counts provided in Table~\ref{tab:model_human_ratio}, \textbf{Claude} emerges as the most effective annotator among the four models, despite its own specific biases. Claude\cite{anthropic2024claude} is the only model that consistently identifies long-tail semantic categories with frequencies comparable to humans. For example, its ratios for International (0.96), Economy (1.06), Sports (1.08), and Sarcasm (0.94) are remarkably close to the human baseline of 1.00. While it tends to over-detect National and Negative content, it avoids the catastrophic label space compression seen in the other models.

From Table~\ref{tab:model_human_ratio} GPT, Gemini, and Grok are most conservative or fallback biased. These models show severe over-representation of fallback labels and extreme under-representation of informative classes. Specifically:
\begin{itemize}
    \item \textbf{Fallback Labels:} These models exhibit \textit{Other} ratios ranging between 1.34 and 1.52.
    \item \textbf{GPT:} Captures only 10\% of Technology and 20\% of Hateful content identified by humans.
    \item \textbf{Gemini:} Is nearly ``blind'' to Sarcasm, identifying only 2 instances where humans found 35 (Ratio: 0.06).
    \item \textbf{Grok:} Exhibits the highest reliance on the \textit{Other} category (441 instances vs. 291 human) and the Neutral sentiment (Ratio: 1.49).
\end{itemize}

\subsection{False Negative Analysis}

We further analyze false negatives, defined as instances where LLMs assign fallback labels (e.g., \textit{No}, \textit{Neutral}, \textit{Other}) while human annotators identify them as semantically meaningful classes. 

\begin{table}[h]
    \centering
    \caption{False Negative (FN) Impact on Nuanced Classes}
    \label{tab:fn_analysis}
    \addtolength{\tabcolsep}{-5pt}
    \small
    \begin{tabular}{lccc}
        \toprule
        \cellcolor{gray!15}\textbf{Class} & \cellcolor{gray!15}\textbf{Human} & \cellcolor{gray!15}\textbf{LLM Avg.} & \cellcolor{gray!15}\textbf{FN Rate (\%)} \\ \midrule
        \cellcolor{red!10}Hateful (Yes)  & 46  & 36.75 & \cellcolor{red!10}79.9\% \\
        \cellcolor{red!10}Sarcasm (Yes)  & 35  & 26.25 & \cellcolor{red!10}75.0\% \\
        Sentiment (+/-)& 184 & 84.75 & 46.1\% \\
        Specialized Cat.* & 84  & 48.25 & 57.4\% \\ \bottomrule
        \multicolumn{4}{l}{\scriptsize *Sum of Economy, Tech, Sports, Education, and International.}
    \end{tabular}
\end{table}

A large proportion of missed cases is observed in sarcasm and hateful content detection, where the FN rate exceeds 75\% (Table \ref{tab:fn_analysis}). This suggests that LLMs exhibit a "conservative bias," systematically failing to detect implicit, context-dependent, or pragmatically encoded expressions which are readily apparent to human readers. The False Negative Rate of 79\% and 75\% respectively for Hateful Content and Sarcasm annotations represents one of the most important results obtained from this experiment. In effect, this result indicates that the use of LLMs for language annotation purposes seems almost blind to the most intricate characteristics of human communication. On top of this, the FN Rate in specialized cases (57.4\%) reinforces the hypothesis that LLMs focus on a generalist interpretation rather than specialized information. The implication of this result for scientists is clear: using LLMs to annotate data would mean to obtain "cleansed" data.


\subsection{Implications for Annotation Reliability}

Analysis suggests that LLM-powered consistency is an “agreement illusion,” where high reliability metrics (e.g., Fleiss’ Kappa) may emerge due to the \textbf{dominance of the fallback label} instead of semantic consistency with human perception.

Thus, a key issue in data engineering becomes \textit{semantic space compression}. Closed-set LLM-based annotations are biased towards the underrepresentation of less frequent categories or subtle occurrences like irony or implicit toxicity. As a result, the usage of these “silver-standard” datasets can lead to models incapable of recognizing rare but essential language signals.

\section{Benchmarking Under Biased Annotations}
\label{sec:bench}
This section evaluates the impact of LLM-induced annotation bias on downstream benchmarking tasks. We analyze how training and evaluation under LLM-generated labels affects model behavior.

\subsection{Experimental Setup}

We evaluate the impact of LLM-induced annotation bias using the following configuration:

\begin{itemize}
    \item \textbf{Data and Supervision}: Models are trained on the full 58K LLM-annotated MultiSoc-4D dataset.
    \item \textbf{Task Setting}: We perform \textit{multi-label classification} across the four analyzed dimensions.
    \item \textbf{Model Selection}: Our benchmark includes a wide array of models:
    \begin{itemize}
        \item \textbf{LLMs}: Qwen2.5 (0.5B-7B)\cite{qwen2024qwen25}, LLaMA3.2 (1B-3B), LLaMA3.1 (8B), TinyLLaMA 1.1B, Aya Expanse 8B, Phi-3/4, Gemma variants and Mistral\cite{jiang2023mistral} variants.
        \item \textbf{Transformers}: XLM-RoBERTa, MuRIL, RemBERT, BanglaBERT, DeBERTa, XLNet, Electra, FinBERT, BanglaT5, and IndicBERT.\cite{wolf2020transformers}
        \item \textbf{Traditional ML}: Linear/Logistic Regression, Ridge/Lasso Classifier, SVM, KNN, Random Forest, Decision Tree, Gradient Boost, AdaBoost, Multinomial/Gaussian NB, LightGBM, XGBoost, and CatBoost.\cite{bishop2006pattern}
    \end{itemize}
    \item \textbf{Hyperparameters}: Training is conducted using PyTorch and Hugging Face with LoRA adaptation. We use a learning rate of $2e-5$, a batch size of 32, and 1 training epoch.
    \item \textbf{Hardware}: All experiments are performed on an NVIDIA RTX 4070 Ti Super GPU.
    \item \textbf{Evaluation Metrics}:We evaluated the performance of several models on this dataset using standard text classification metrics, including \textbf{Accuracy}, \textbf{Precision}, \textbf{Recall}, and the \textbf{F1-score}. While these metrics provide a broad overview of model efficacy, our benchmark study primarily focuses on the \textbf{Macro-F1 score}. This choice was made to rigorously analyze class-wise performance and mitigate the impact of class imbalance (bias). By averaging the F1-scores of each class independently, the Macro-F1 metric ensures that the model's ability to identify minority classes is treated with equal importance to the majority classes. The Macro-F1 score is defined as the arithmetic mean of the individual F1-scores for all classes, as shown in Equation \ref{eq:macrof1}:

        \begin{equation}
        \label{eq:macrof1}
        \text{Macro-F1} = \frac{1}{N} \sum_{i=1}^{N} F1_i
        \end{equation}
    where $N$ represents the total number of classes and $F1_i$ is the F1-score for the $i$-th class, calculated as:
    \begin{equation}
    F1_i = 2 \times \frac{\text{Precision}_i \times \text{Recall}_i}{\text{Precision}_i + \text{Recall}_i}
    \end{equation}
\end{itemize}


\subsection{Model Performance}

\begin{table}[H]
    \centering
    \footnotesize 
    \setlength{\tabcolsep}{2pt} 
    \caption{Performance Comparison Across Model Architectures}
    \label{tab:model_comparison_multirow}
    \begin{tabular}{llccc}
        \toprule
        \cellcolor{gray!15}\textbf{Model Type} & \cellcolor{gray!15}\textbf{Dim.} & \cellcolor{gray!15}\textbf{Model} & \cellcolor{gray!15}\textbf{Acc.}  & \cellcolor{green!15}\textbf{M-F1} \\ \midrule
        
        \multirow{4}{*}{\shortstack[l]{Instruction-\\Tuned LLMs}} 
        & Category  & Qwen2.5 0.5B & 0.77 & \cellcolor{green!15}0.73 \\
        & Sentiment & Mistral Nemo & 0.78 & \cellcolor{green!15}0.68 \\
        & Hateful   & Mistral Nemo & 0.96 & \cellcolor{green!15}0.71 \\
        & Sarcasm   & Mistral 7B   & 0.96 & \cellcolor{green!15}0.54 \\ \midrule
        
        \multirow{4}{*}{\shortstack[l]{Multilingual\\Models}} 
        & Category  & RemBERT      & 0.78 & \cellcolor{green!15}0.73 \\
        & Sentiment & IndicBERT    & 0.79 & \cellcolor{green!15}0.70 \\
        & Hateful   & IndicBERT    & 0.97 & \cellcolor{green!15}0.74 \\
        & Sarcasm   & IndicBERT    & 0.96 & \cellcolor{green!15}0.58 \\ \midrule
        
        \multirow{4}{*}{\shortstack[l]{Monolingual\\Models}} 
        & Category  & BanglaBERT   & 0.75 & \cellcolor{green!15}0.71 \\
        & Sentiment & BanglaBERT   & 0.79 & \cellcolor{green!15}0.68 \\
        & Hateful   & BanglaBERT   & 0.97 & \cellcolor{green!15}0.77 \\
        & Sarcasm   & BanglaBERT   & 0.96 & \cellcolor{green!15}0.59 \\ \bottomrule
    \end{tabular}
\end{table}

From the Table~\ref{tab:model_comparison_multirow}, it is evident that Monolingual and Multilingual transformers (BanglaBERT and IndicBERT) perform better compared to Instruction-Tuned LLMs in detecting subtle aspects of languages. The LLMs exhibit a very high accuracy, but in terms of the Macro F1 score, they perform poorly especially on Sarcasm. It is likely that these models cannot detect the minority classes properly due to their complexity. On the other hand, the best-performing model in the detection of critical aspects such as Hatefulness is the BanglaBERT with a Macro F1 score of 0.77.


\subsection{Human-Aligned Evaluation}

To assess the impact of label space compression on downstream model performance, we evaluate our top-performing models on the human-annotated gold subset. Given that Claude exhibited the highest distributional alignment with human labels in our previous analysis, we compare models trained on Claude-annotated data against the same models evaluated on human gold-standard labels.  The results, detailed in Table \ref{tab:human_aligned_eval}, show that models evaluated against human annotations generally exhibit a degradation in performance—particularly in accuracy—compared to their evaluation on LLM-generated labels. This confirms that training on collapsed label distributions reduces model sensitivity to minority semantic phenomena and reinforces conservative prediction behaviors that diverge from actual human judgment. 

\begin{table}[h!]
\centering
\caption{Model Performance: Claude-Annotated vs. Human Gold Standard (N=500)}
\label{tab:human_aligned_eval}
\addtolength{\tabcolsep}{-3pt}
\small
\begin{tabular}{@{}lllcc@{}}
\toprule
\cellcolor{gray!15}\textbf{Dimension} & \cellcolor{gray!15}\textbf{Model} & \cellcolor{gray!15}\textbf{Eval. Set} & \cellcolor{gray!15}\textbf{Acc.} & \cellcolor{green!15}\textbf{M-F1} \\ \midrule
\multirow{6}{*}{Category} & \multirow{2}{*}{Qwen2.5 0.5B} & Claude & 0.38 & \cellcolor{green!15}0.18 \\
 & & Human & 0.54 & \cellcolor{green!15}0.16 \\ 
 &\multirow{2}{*}{RemBERT} & Claude & 0.39 & \cellcolor{green!15}0.09 \\
 & & Human & 0.58 & \cellcolor{green!15}0.09 \\ 
 &\multirow{2}{*}{BanglaBERT} & Claude & 0.37 & \cellcolor{green!15}0.09 \\
 & & Human & 0.58 & \cellcolor{green!15}0.09 \\\midrule
\multirow{6}{*}{Sentiment} & \multirow{2}{*}{Mistral Nemo} & Claude & 0.43 & \cellcolor{green!15}0.40 \\
 & & Human & 0.59 & \cellcolor{green!15}0.42 \\ 
 & \multirow{2}{*}{IndicBERT} & Claude & 0.50 & \cellcolor{green!15}0.24 \\
 & & Human & 0.63 & \cellcolor{green!15}0.26 \\
 & \multirow{2}{*}{BanglaBERT} & Claude & 0.52 & \cellcolor{green!15}0.32 \\
 & & Human & 0.63 & \cellcolor{green!15}0.26 \\\midrule
\multirow{6}{*}{Hateful} & \multirow{2}{*}{Mistral Nemo} & Claude & 0.95 & \cellcolor{green!15}0.49 \\
 & & Human & 0.89 & \cellcolor{green!15}0.49 \\ 
 & \multirow{2}{*}{IndicBERT} & Claude & 0.94 & \cellcolor{green!15}0.48 \\
 & & Human & 0.93 & \cellcolor{green!15}0.48 \\ 
 & \multirow{2}{*}{BanglaBERT} & Claude & 0.94 & \cellcolor{green!15}0.48 \\
 & & Human & 0.93 & \cellcolor{green!15}0.48 \\ \midrule
\multirow{6}{*}{Sarcasm} & \multirow{2}{*}{Mistral3 7B} & Claude & 0.92 & \cellcolor{green!15}0.48 \\
 & & Human & 0.93 & \cellcolor{green!15}0.48 \\ 
 & \multirow{2}{*}{IndicBERT} & Claude & 0.97 & \cellcolor{green!15}0.49 \\
 & & Human & 0.94 & \cellcolor{green!15}0.48 \\ 
 & \multirow{2}{*}{BanglaBERT} & Claude & 0.97 & \cellcolor{green!15}0.49 \\
 & & Human & 0.94 & \cellcolor{green!15}0.48 \\ \bottomrule
\end{tabular}
\end{table}
The evaluation reveals a critical performance gap between LLM-derived and human-derived performance metrics. In the Category and Sentiment dimensions, the accuracy is significantly higher when evaluated against humans (reaching 0.54 and 0.63 respectively), suggesting that the models' "reasoning" aligns more closely with human labels than the sanitized Claude annotations.  However, in the high-stakes dimensions of Hateful and Sarcasm, the performance remains nearly identical across both datasets, with high accuracy ($\approx 0.92$) and low Macro F1 ($\approx 0.44$). This indicates that even the best-performing models (Mistral Nemo and Mistral3 7B) are struggling with the "Accuracy-F1 Paradox". They successfully predict the majority label ("No") but fail to reliably capture the nuance identified by humans, proving that even a "human-aligned" LLM like Claude still provides a training signal that is insufficient for mastering complex, context-dependent semantic phenomena.  


\subsection{Bias Sensitivity Across Models}

\begin{figure}[t]
    \centering
    \includegraphics[width=\linewidth]{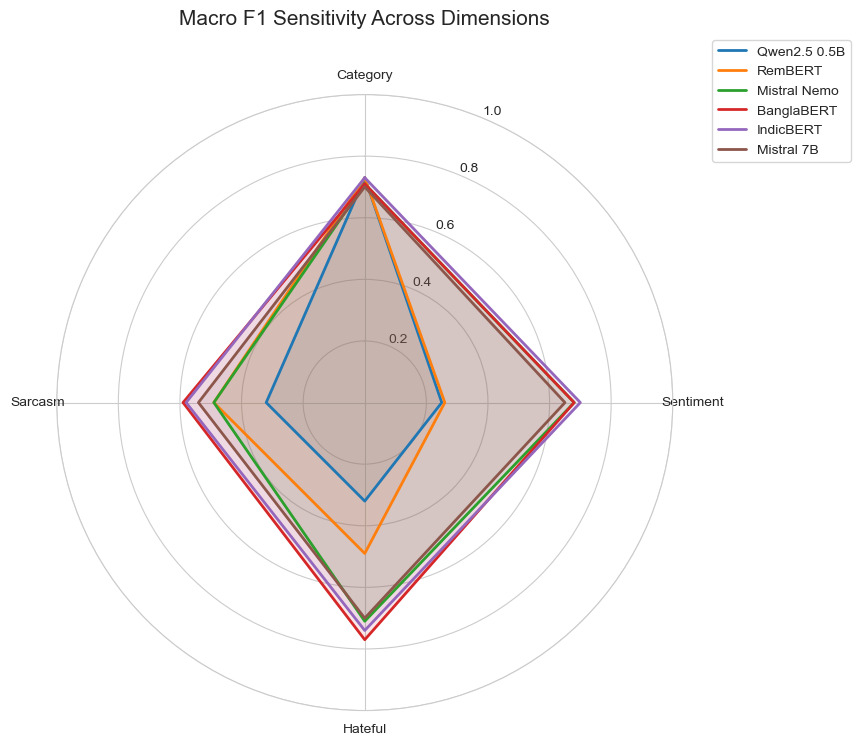}
    \caption{Macro-F1 sensitivity accross models}
    \label{fig:bias_sensitivity_radar}
\end{figure}

Finally, we analyze the response of diverse model architectures to systematic annotation bias. As visualized in Figure \ref{fig:bias_sensitivity_radar}, we find that all models—ranging from small-scale Transformers to large instruction-tuned LLMs—exhibit nearly identical sensitivity patterns. While models achieve high accuracy on fallback-dominated tasks, there is a universal failure to generalize to human-identified semantic classes.
This uniformity indicates that annotation bias is an architectural-agnostic phenomenon; it propagates through the training pipeline and systematically restricts downstream model behavior, regardless of the specific model choice.

\section{Discussion}
\label{sec:dis}
\subsection{Label Ambiguity Analysis}
However, the difference in the human vs. LLM tagging can be attributed to the fact that social media language is inherently ambiguous. While humans use contextual knowledge to decipher the meaning in ambiguous phrases, LLMs find it difficult to comprehend pragmatically coded expressions. The analysis we have conducted reveals that the response of an LLM to ambiguous prompts is that of a ``majority-class consensus.''

\subsection{Other-Class Overloading}
One interesting result from the MultiSoc-4D dataset was the "overloading" of the "Other" category. Although humans carefully classified documents into different categories including Technology, International, and Education, LLMs such as GPT and Grok categorized far too many samples into the "Other" category. The label space compression essentially filters out the long-tail data variety within the dataset.

\subsection{Neutral vs Non-Sentiment Cases}
We notice an important conflation between "Neutral" sentiment and "Non-Sentiment" instances. LLMs display a ``conservative bias'' and are more likely to classify the data as "neutral," even when there are triggers for sentiments embedded in the data. The accuracy-F1 paradox of the Sentiment axis is supported by the observation that the high accuracy of the Sentiment axis is due to the prevalence of the Neutral label.

\subsection{Error Analysis}
From our error analysis, we conclude that False Negatives for Sarcasm and Hate Speech are the main reasons behind the failure of the model. The relatively high False Negative rate (over 75\% in some models) implies that large language models tend to underestimate implicit toxicity, which results in ``sterilizing" those data samples that are most socially linguistically complex.

\section{Limitations and Future Work}
\label{sec:lim}
\subsection{Closed Label Schema Limitation}
The limitations of the present study are confined to its closed-set labeling methodology, where the models are forced to select labels from a fixed set of labels. Such limitations might be causing the ``dominance of fallback labels'' phenomenon because the models cannot indicate their uncertainty or propose other semantic labels.

\subsection{Instruction-Induced Bias}
The behavior of the instruction-tuned models employed for annotations (such as LLaMA, Qwen, Phi) is strongly affected by the safety alignment within them. It appears that such “safety-first” training affects the models’ unwillingness to classify content into Hateful and Sarcastic categories, thus causing the underrepresentation of the two minority classes in the silver standard data set.

\subsection{Future Direction: Open-Set Labeling (NaN)}
For tackling the issue of label space reduction, future research will consider open set labeling with the use of "NaN" labels or labels denoted by "Uncertain". The objective is to enable models to provide natural language explanations or indicate when a label does not fit the schema.

\section{Ethical Considerations}
\label{sec:ec}
\subsection{Data Privacy}
The MultiSoc-4D dataset is derived from publicly available social media text. To protect user privacy, all personally identifiable information (PII) has been removed, and the research adheres to standardized academic protocols for large-scale data analysis.

\subsection{Content Sensitivity}
This research involves the analysis of Hateful Content and Sarcasm. While essential for developing robust safety filters, exposure to such content during the human annotation process was managed with care to minimize psychological impact on the annotators.

\subsection{Bias and Fairness}
We acknowledge that the LLMs used for benchmarking—such as Claude, Gemini, and GPT—contain inherent biases from their pre-training data. By comparing these against human gold standards, this study explicitly aims to highlight and quantify these biases to foster more fair and transparent automated annotation pipelines.

\section{Conclusion}
\label{sec:con}
MultiSoc-4D presents a Bengali dataset spanning multiple platforms where it becomes evident that label collapse is a system failure for closed-set LLM annotation. Our analysis of four state-of-the-art LLMs on a common validation set proves that this is due to the instruction paradigm and not because of any model-specific shortcomings. We measure this phenomenon using benchmarking on a human reference dataset of 500 instances. LLMs are able to catch less than one-quarter of the offensive and sarcastic posts that humans mark as such, having false negatives at 79\% and 75\%, respectively. However, the high agreement across models seen throughout dimensions can be dismissed as a misleading statistic due to their tendency towards convergence on fallback labels rather than any actual mutual semantic comprehension. As illustrated by Fleiss' $\kappa$ score of  $\approx -0.001$ for sarcasm and nearly 96\% agreement on "No," such an illusion becomes statistically concrete. Cross-benchmarking down the pipeline with more than 40 models, including traditional ML, transformer-based models, and instruction-tuned models, reveals that the bias is not remedied during training. Instead, it perpetuates. These models acquire the same conservative tendencies, achieving high precision on majority fallback classes yet failing on the minority classes crucial for downstream tasks such as content moderation and hate speech detection. MultiSoc-4D is released as a diagnostic tool that explicitly measures, rather than conceals, annotation bias. While human reference data quantifies the gap, it is not a final correction.





\bibliography{custom}
\newpage
\clearpage
\appendix

\section{Annotation Guidelines}
\label{sec : AG}
\subsection*{Instruction}
This section details the instructions provided to annotators for the Bangla Comment Dataset to ensure consistency and high-quality labeling.

\subsubsection*{Annotation Schema}
Each comment is evaluated across four distinct dimensions:
\begin{itemize}[noitemsep, topsep=2pt]
    \item \textbf{Category:} (8 classes) International, National, Entertainment, Education, Sports, Technology, Economy, Other.
    \item \textbf{Sentiment:} (3 classes) Positive, Negative, Neutral.
    \item \textbf{Hatefulness:} (Binary) Yes, No.
    \item \textbf{Sarcasm:} (Binary) Yes, No.
\end{itemize}

\subsubsection*{Standardized Examples}
The following table provides representative examples for various annotation dimensions.

\begin{table}[h]
    \small
    \centering
    \begin{tabularx}{\linewidth}{lXc}
        \toprule
        \cellcolor{gray!15}\textbf{Dimension} & \cellcolor{gray!15}\textbf{Bangla Comment} & \cellcolor{gray!15}\textbf{Label} \\ \midrule
        Category & \textbengali{গাজায় যুদ্ধ পরিস্থিতি খারাপ হচ্ছে} & International \\
        Category & \textbengali{বাংলাদেশে নতুন বাজেট ঘোষণা} & National \\
        Category & \textbengali{বাংলাদেশ ম্যাচ জিতে গেছে} & Sports \\ \midrule
        Sentiment & \textbengali{দারুণ খেলেছে বাংলাদেশ} & Positive \\
        Sentiment & \textbengali{এই সিদ্ধান্ত খুব খারাপ} & Negative \\
        Sentiment & \textbengali{আজ স্কুল বন্ধ} & Neutral \\ \midrule
        Hateful & \textbengali{ওদের জাতটাই খারাপ} & Yes \\
        Sarcasm & \textbengali{ওয়াও! কি দারুণ ব্যর্থতা!} & Yes \\
        \bottomrule
    \end{tabularx}
\end{table}

\subsubsection*{Full Annotation Samples}
Annotators are required to populate the fields in the sequence: \textit{Category $\rightarrow$ Sentiment $\rightarrow$ Hateful $\rightarrow$ Sarcasm}.

\begin{table}[h]
    \small
    \centering
    \begin{tabularx}{\linewidth}{Xcccc}
        \toprule
        \cellcolor{gray!15}\textbf{Comment} & \cellcolor{gray!15}\textbf{Category} & \cellcolor{gray!15}\textbf{Sent.} & \cellcolor{gray!15}\textbf{Hate} & \cellcolor{gray!15}\textbf{Sarc} \\ \midrule
        \textbengali{বাংলাদেশ ম্যাচ জিতে গেছে} & Sports & Pos. & No & No \\
        \textbengali{তুই useless} & Other & Neg. & Yes & No \\
        \textbengali{ওয়াও! আবার ফেল করলি} & Education & Neg. & No & Yes \\
        \textbengali{আগামীকাল পরীক্ষা} & Education & Neu. & No & No \\
        \bottomrule
    \end{tabularx}
\end{table}

\subsubsection*{Annotation Rules \& Conflict Resolution}
\begin{enumerate}[label=\arabic*., leftmargin=*, nosep]
    \item \textbf{Default Assignment (Confusion Rule):} In cases of high ambiguity or lack of context, annotators must default to: \textit{Other, Neutral, No, No}.
    \item \textbf{Linguistic Nuance:} Criticism of an idea is labeled as \textbf{Hateful: No}, whereas attacks on identity or personhood are labeled as \textbf{Hateful: Yes}.
    \item \textbf{Sarcasm Identification:} Sarcasm is only labeled \textbf{Yes} if the intended meaning is the opposite of the literal text (irony).
\end{enumerate}
\subsection*{Annotation Prompt}

\begin{tcolorbox}[colback=gray!5,colframe=black,title=LLM Annotation Prompt]
\small

\textbf{Role:} Act as a data annotator specializing in Bengali text classifier.

\vspace{0.3em}
\textbf{Input:}
You are provided with:
\begin{itemize}
    \item A CSV file containing user comments.
    \item An instruction document (Instruction.docx) defining labeling criteria.
\end{itemize}

\vspace{0.3em}
\textbf{Task:}
\begin{itemize}
    \item Parse the text in the \texttt{comments} column.
    \item Annotate each entry following the hierarchy and definitions in \texttt{Instruction.docx}.
    \item Classify data into standardized labels defined in the instruction file.
    \item Prioritize neutral labeling where appropriate.
\end{itemize}

\vspace{0.3em}
\textbf{Output Requirements:}
\begin{itemize}
    \item Return a CSV file containing only:
    \begin{itemize}
        \item Original comments
        \item Generated labels
    \end{itemize}
    \item Ensure no extra commentary or metadata is included.
    \item Assign your name in the \texttt{Annotator Name} field.
\end{itemize}

\vspace{0.3em}
\textbf{Final Instruction:}
Provide the output as a downloadable CSV file only.

\end{tcolorbox}

\section{Dataset Examples}
\label{sec : dex}
Datasets are diverse but the annotations seem noisy. The sample of the dataset is shown in the Table~\ref{tab:dataset_samples}.

\begin{table*}[ht!]
    \centering
    \caption{Sample Annotated Dataset}
    \label{tab:dataset_samples}
    \small
    \renewcommand{\arraystretch}{1} 
    \begin{tabular}{lp{6cm}cccc}
        \toprule
        \cellcolor{gray!15}\textbf{Source} & \cellcolor{gray!15}\textbf{Comment (Bangla)} & \cellcolor{gray!15}\textbf{Cat.} & \cellcolor{gray!15}\textbf{Sent.} & \cellcolor{gray!15}\textbf{Hate} & \cellcolor{gray!15}\textbf{Sarc.} \\ 
        \midrule
        Instagram & \textbengali{ভারতে করোনায় সর্বোচ্চ মৃত্যু আর আক্রান্তের নতুন রেকর্ড} & International & Negative & No & No \\
        YouTube & \textbengali{তুই তো এখন বাংলাদেশের নতুন মাগি} & National & Negative & Yes & No \\
        TikTok & \textbengali{আল্লাহ মুসলিম রাষ্ট্র নায়কদের এক হয়ে অন্য মুসলমান ভাইদের নিরাপত্তার জন্য এক হওয়ার তৌফিক দান করুন।} & Other & Neutral & No & No \\
        Facebook & \textbengali{চেক কেটে ভিডিও পাঠালেও টাকা ফেরত পাচ্ছেন না ইভ্যালির গ্রাহকরা} & Other & Neutral & No & Yes \\
        Twitter & \textbengali{"আমার লাভ লস নাই,, আমার জীবনটায় লস"} & Other & Negative & No & No \\
        Facebook & \textbengali{করোনা টিকা নিতে কেন্দ্রে বাড়ছে ভিড়} & Other & Neutral & No & No \\
        Likee & \textbengali{বেসরকারি হাসপাতাল খোলা রাখার আহ্বান স্বাস্থ্যমন্ত্রীর} & National & Neutral & No & No \\
        Facebook & \textbengali{সাকিবই শেষ ভরসা} & Sports & Neutral & No & No \\
        \bottomrule
    \end{tabular}
\end{table*}

\section{Additional Dataset Visualization}
\label{sec: DE}
\begin{figure}[t]
    \centering
    \includegraphics[width=\linewidth]{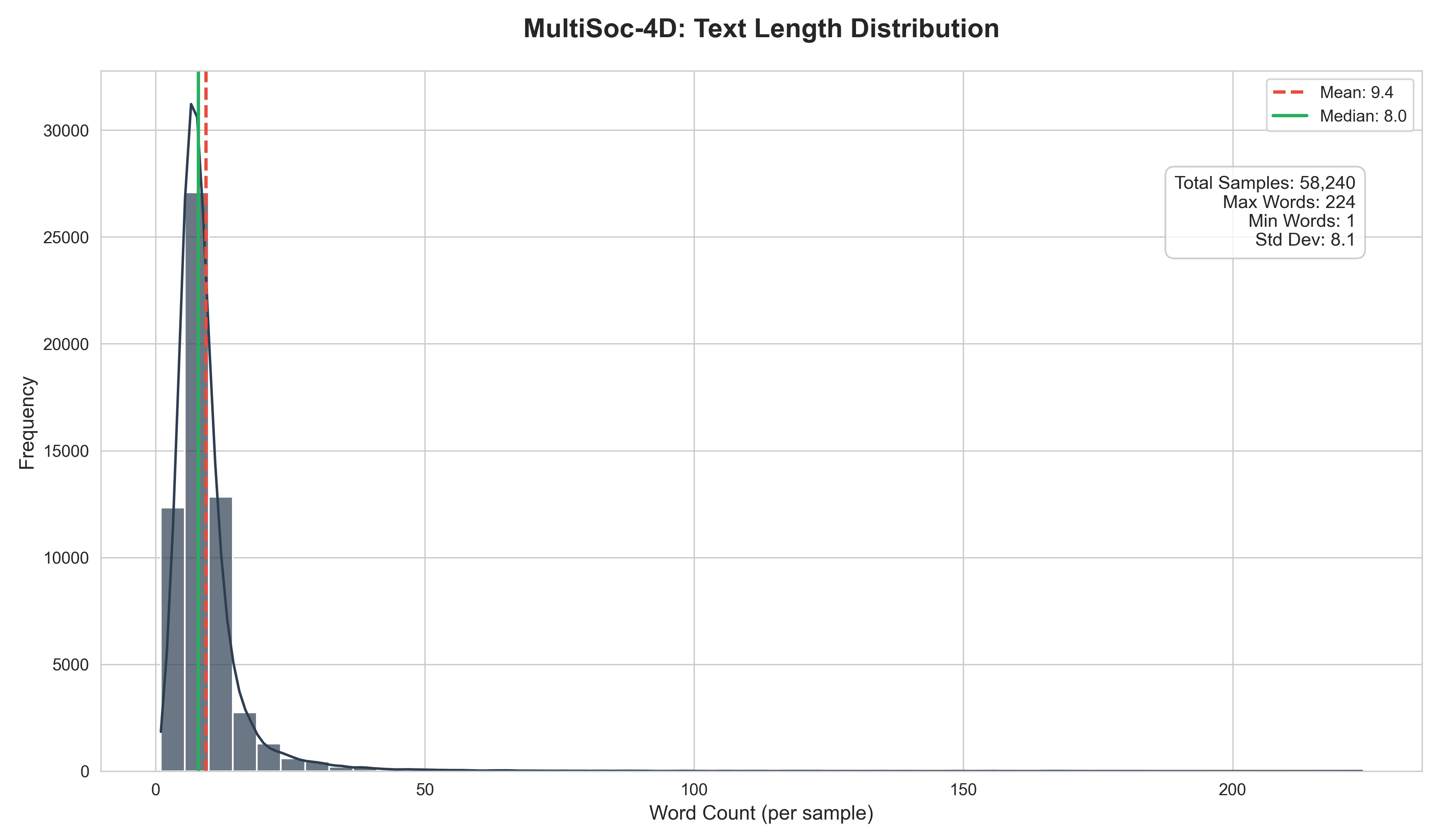}
    \caption{The dataset contains both short and long text samples, ranging from brief comments to longer conversational statements which is visually shown in. This variation reflects realistic usage patterns and introduces additional complexity for annotation.}
    \label{fig:distribution11}
\end{figure}

The text visualization presented in Figure~\ref{fig:distribution11} which shows the range of the text present in the dataset as "Comment".

\section{Extended Results Tables}
The benchmarking was categorized by 4 types of model-pipeline selection. Those are : 
\begin{enumerate}
    \item Instruction-Tuned Large Language Models
    \item Multi-Lingual Tranformers (LLM)
    \item Mono-Lingual Transformers (LLM)
    \item Traditional Machine Learning Model
\end{enumerate}

The performance benchmark of the instruction-tuned llms are shown in the Table~\ref{tab:model_performance1}. For transformer based mono and multi-lingual models performance are presented in Table~\ref{tab:model_performance2}. And the Table~\ref{tab:model_performance3} shows the benchmarking of traditional machine learning models.

\begin{table}[h!]
    \centering
    \caption{Performance Comparison of Instruction-Tuned Models Across Four Dimensions}
    \label{tab:model_performance1}
    \small
    \begin{tabular}{llcc}
        \toprule
        \cellcolor{gray!15}\textbf{Dim.} & \cellcolor{gray!15}\textbf{Model} & \cellcolor{gray!15}\textbf{Acc.} & \cellcolor{gray!15}\textbf{M-F1} \\ \midrule
        \multirow{15}{*}{Category} & Qwen2.5 0.5B & 0.77 & 0.73 \\
                                   & Qwen2.5 1.5B & 0.78 & 0.67 \\
                                   & Qwen2.5 3B &0.75  &0.68\\
                                   & Qwen2.5 7B &0.75  &0.68\\
                                   
                                   & LLaMA3.1 8B  & 0.75 & 0.68 \\
                                   & LLaMA3.2 1B  & 0.76 & 0.69 \\
                                   & LLaMA3.2 3B  & 0.76 & 0.70 \\
                                   & TinyLlama 1.1B  & 0.73 & 0.62 \\
                                   & Phi-3        & 0.74 & 0.65 \\
                                   & Phi-4        & 0.74 & 0.67 \\
                                   & Gemma 2B    & 0.16 & 0.10 \\
                                   & Gemma2 2B    & 0.51 & 0.35 \\
                                   
                                   & Mistral 7B &0.76 &0.70\\ 
                                   & Mistral Nemo & 0.77 & 0.71 \\ 
                                   &Aya Expense 8B &0.05 &0.01\\\midrule
                                   
        \multirow{15}{*}{Sentiment} 
        & Qwen2.5 0.5B   & 0.61 & 0.25 \\
        & Qwen2.5 1.5B & 0.78 & 0.44 \\
                                    
                                    & Qwen2.5 3B   & 0.77 & 0.64 \\
                                    & Qwen2.5 7B   & 0.77 & 0.64 \\
                                    & Llama3.1 8B & 0.75  &0.61\\ 
                                    & LLaMA3.2 1B  & 0.78 & 0.66 \\
                                    & LLaMA3.2 3B  & 0.78 & 0.65 \\
                                    &TinyLlama 1.1B  &0.75  &0.59\\
                                    & Phi-3        & 0.75 & 0.62 \\
                                    & Phi-4        & 0.75 & 0.59 \\
                                    & Gemma 2B     & 0.44 & 0.33 \\
                                    & Gemma2 2B    & 0.34 & 0.35 \\
                                    & Mistral 7B & 0.77  &0.65\\
                                    & Mistral Nemo & 0.78 & 0.68 \\ 
                                    &Aya Expense 8B &0.14 &0.08\\
                                    \midrule
        \multirow{15}{*}{Hateful} 
        & Qwen2.5 0.5B   & 0.94 & 0.32 \\
        
        & Qwen2.5 1.5B & 0.97 & 0.46 \\
                                    & Qwen2.5 3B   & 0.96 & 0.68 \\
                                    & Qwen2.5 7B   & 0.96 & 0.70 \\
                                    & LLaMA3.1 8B  & 0.96 & 0.67 \\
                                    & LLaMA3.2 1B  & 0.96 & 0.67 \\
                                    & LLaMA3.2 3B  & 0.96 & 0.69 \\
                                    & TinyLlama 1.1B  &0.96  &0.67\\
                                    & Phi-3        & 0.96 & 0.67 \\
                                    & Phi-4        & 0.96 & 0.66 \\
                                    & Gemma 2B    & 0.09 & 0.09 \\
                                    & Gemma2 2B    & 0.17 & 0.16 \\
                                    & Mistral 7B &0.96  &0.70\\
                                    & Mistral Nemo & 0.96 & 0.71 \\
                                    & Aya Expense 8B  &0.96  &0.49 \\
                                    \midrule

        \multirow{15}{*}{Sarcasm}    
        & Qwen2.5 0.5B    & 0.94 & 0.32 \\
        & Qwen2.5 1.5B    & 0.92 & 0.32 \\
        & Qwen2.5 3B   & 0.96 & 0.49 \\
                                    & Qwen2.5 7B   & 0.96 & 0.49 \\

                                    & Llama3.1 8B    & 0.96 & 0.49 \\
                                    & LLaMA3.2 1B  & 0.96 & 0.51 \\
                                    & Llama3.2 3B    & 0.96 & 0.50 \\
                                    & TinyLlama 1.1B    & 0.96 & 0.49 \\
                                    & Phi-3        & 0.96 & 0.51 \\
                                    & Phi-4   & 0.96 & 0.49 \\
                                    & Gemma 2B    & 0.08 & 0.08 \\
                                    & Gemma2 2B    & 0.17 & 0.16 \\
                                    & Mistral3 7B  & 0.96 & 0.54 \\ 
                                    & Mistral Nemo  & 0.96  &0.49 \\
                                    & Aya Expense 8B &0.94  &0.49\\
                                    \bottomrule
    \end{tabular}
\end{table}

\begin{table}[h!]
    \centering
    \caption{Performance Comparison of Traditional Machine Learning Models Across Four Dimensions}
    \label{tab:model_performance3}
    \small
    \begin{tabular}{llcc}
        \toprule
        \cellcolor{gray!15}\textbf{Dim.} & \cellcolor{gray!15}\textbf{Model} & \cellcolor{gray!15}\textbf{Acc.} & \cellcolor{gray!15}\textbf{M-F1} \\ \midrule
        \multirow{15}{*}{Category} & Liner Regression & 0.11 & 0.17 \\
        & Logestic Regression & 0.64 & 0.39 \\
        & Ridge Classifier &0.64  &0.39\\
        & Lasso Classifier &0.63  &0.33\\
        & SVM  & 0.64 & 0.38 \\
        & KNN  & 0.56 & 0.31 \\
        & Decision Tree  & 0.55 & 0.35 \\
        & Random Forest  & 0.63 & 0.38 \\
        & Gradient Boost       & 0.64 & 0.38 \\
        & AdaBoost       & 0.58 & 0.17\\
        & Multinomial NB  & 0.59 & 0.21 \\
        & Gaussian NB  & 0.09 & 0.11 \\
        & XGBoost  & 0.64 & 0.40 \\
        & LightGBM       & 0.65 & 0.41 \\
        & CatBoost       & 0.06 & 0.01\\\midrule
                                   
        \multirow{15}{*}{Sentiment}& 
        Liner Regression & 0.52 & 0.41 \\
        & Logestic Regression & 0.71 & 0.49 \\
        & Ridge Classifier &0.71  &0.49\\
        & Lasso Classifier &0.71  &0.46\\
        & SVM  & 0.72 & 0.47 \\
        & KNN  & 0.63 & 0.46 \\
        & Decision Tree  & 0.62 & 0.48 \\
        & Random Forest  & 0.71 & 0.49 \\
        & Gradient Boost       & 0.71 & 0.47 \\
        & AdaBoost       & 0.68 & 0.36\\
        & Multinomial NB  & 0.69 & 0.41 \\
        & Gaussian NB  & 0.23 & 0.21 \\
        & XGBoost  & 0.72 & 0.50 \\
        & LightGBM       & 0.72 & 0.50 \\
        & CatBoost       & 0.19 & 0.11\\\midrule
        
        \multirow{15}{*}{Hateful} &
        Liner Regression & 0.95 & 0.51 \\
        & Logestic Regression & 0.96 & 0.52 \\
        & Ridge Classifier &0.95  &0.51\\
        & Lasso Classifier &0.95  &0.50\\
        & SVM  & 0.96 & 0.51 \\
        & KNN  & 0.95 & 0.53 \\
        & Decision Tree  & 0.93 & 0.57 \\
        & Random Forest  & 0.95 & 0.52 \\
        & Gradient Boost       & 0.96 & 0.53 \\
        & AdaBoost       & 0.95 & 0.51\\
        & Multinomial NB  & 0.95 & 0.49 \\
        & Gaussian NB  & 0.26 & 0.24 \\
        & XGBoost  & 0.96 & 0.55 \\
        & LightGBM       & 0.96 & 0.54 \\
        & CatBoost       & 0.96 & 0.53\\\midrule

        \multirow{15}{*}{Sarcasm}   &
        Liner Regression & 0.96 & 0.49 \\
        & Logestic Regression & 0.96 & 0.50 \\
        & Ridge Classifier &0.96  &0.49\\
        & Lasso Classifier &0.96  &0.51\\
        & SVM  & 0.96 & 0.49 \\
        & KNN  & 0.96 & 0.54 \\
        & Decision Tree  & 0.95 & 0.59 \\
        & Random Forest  & 0.96 & 0.51 \\
        & Gradient Boost       & 0.96 & 0.53 \\
        & AdaBoost       & 0.96 & 0.49\\
        & Multinomial NB  & 0.96 & 0.49 \\
        & Gaussian NB  & 0.32 & 0.27 \\
        & XGBoost  & 0.96 & 0.57 \\
        & LightGBM       & 0.96 & 0.53 \\
        & CatBoost       & 0.96 & 0.51\\
        \bottomrule
    \end{tabular}
\end{table}

\begin{table}[h!]
    \centering
    \caption{Performance Comparison of MLM Models Across Four Dimensions}
    \label{tab:model_performance2}
    \small
    \begin{tabular}{llcc}
        \toprule
        \cellcolor{gray!15}\textbf{Dim.} & \cellcolor{gray!15}\textbf{Model} & \cellcolor{gray!15}\textbf{Acc.} & \cellcolor{gray!15}\textbf{M-F1} \\ \midrule
        \multirow{10}{*}{Category} & XLM-RoBERTa & 0.52 & 0.09\\
                                   & Muril & 0.66 & 0.45 \\
                                   & BanglaBERT &0.75  &0.71\\
                                   & IndictBERT &0.77  &0.73\\
                                   & BanglaT5  & 0.06 & 0.01 \\
                                   & Electra  & 0.52 & 0.09 \\
                                   & RemBERT  & 0.78 & 0.73 \\
                                   & FinBERT  & 0.72 & 0.48 \\
                                   & DeBERTa       & 0.70 & 0.51 \\
                                   & XLNet       & 0.52 & 0.09\\\midrule
                                   
        \multirow{10}{*}{Sentiment} 
        & XLM-RoBERTa   & 0.66 & 0.26 \\
        & Muril & 0.78 & 0.64 \\
                                    
                                    & BanglaBERT   & 0.79 & 0.68 \\
                                    & IndicBERT   & 0.79 & 0.70 \\
                                    & BanglaT5 & 0.16  &0.09\\ 
                                    & Electra  & 0.66 & 0.26 \\
                                    & RemBERT  & 0.66 & 0.26 \\
                                    &FinBERT  &0.69  &0.62\\
                                    & DeBERTa        & 0.77 & 0.66 \\
                                    & XLNet        & 0.66 & 0.26 \\
                                    \midrule
        \multirow{10}{*}{Hateful} 
        & XLM-RoBERTa   & 0.95 & 0.49 \\
        & Muril & 0.96 & 0.71 \\
                                    
                                    & BanglaBERT   & 0.97 & 0.77 \\
                                    & IndicBERT   & 0.97 & 0.74 \\
                                    & BanglaT5 & 0.95  &0.49\\ 
                                    & Electra  & 0.95 & 0.49 \\
                                    & RemBERT  & 0.95 & 0.49 \\
                                    &FinBERT  &0.95  &0.54\\
                                    & DeBERTa        & 0.96 & 0.68 \\
                                    & XLNet        & 0.95 & 0.49 \\
                                    \midrule

        \multirow{10}{*}{Sarcasm}    
        & XLM-RoBERTa   & 0.96 & 0.49 \\
        & Muril & 0.96 & 0.49 \\
                                    
                                    & BanglaBERT   & 0.96 & 0.59 \\
                                    & IndicBERT   & 0.96 & 0.58 \\
                                    & BanglaT5 & 0.96  &0.49\\ 
                                    & Electra  & 0.96 & 0.49 \\
                                    & RemBERT  & 0.96 & 0.49 \\
                                    &FinBERT  &0.96  &0.59\\
                                    & DeBERTa        & 0.96 & 0.49 \\
                                    & XLNet        & 0.96 & 0.49 \\
                                    \bottomrule
    \end{tabular}
\end{table}

\section{Results Visualization}
\begin{figure}
    \centering
    \includegraphics[width=\linewidth]{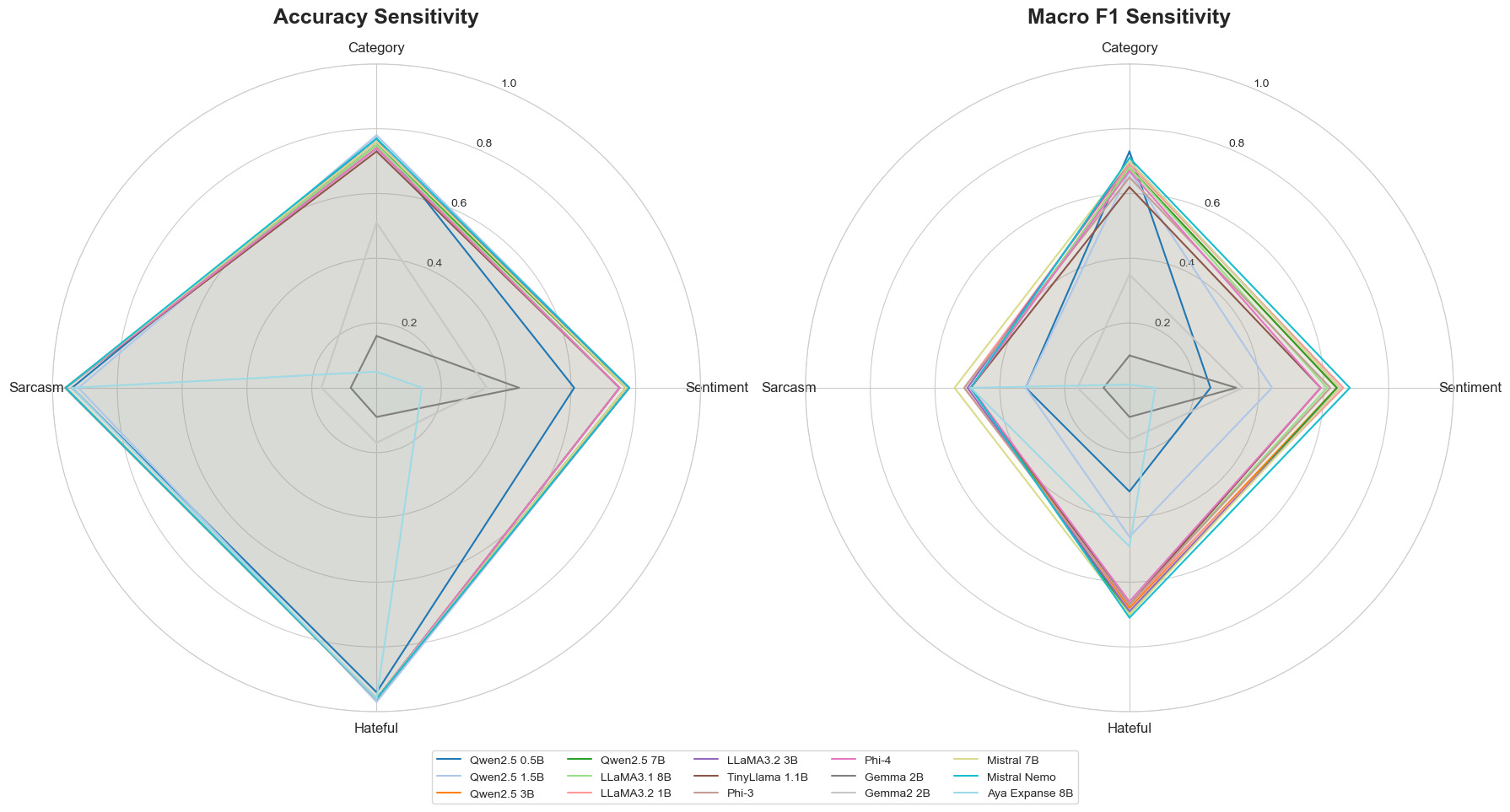}
    \caption{Comparative Radar Charts illustrating Instruction-Tuned Models Sensitivity across Accuracy and Macro F1 dimensions.}
    \label{fig:llmsen}
\end{figure}
\begin{figure}
    \centering
    \includegraphics[width=\linewidth]{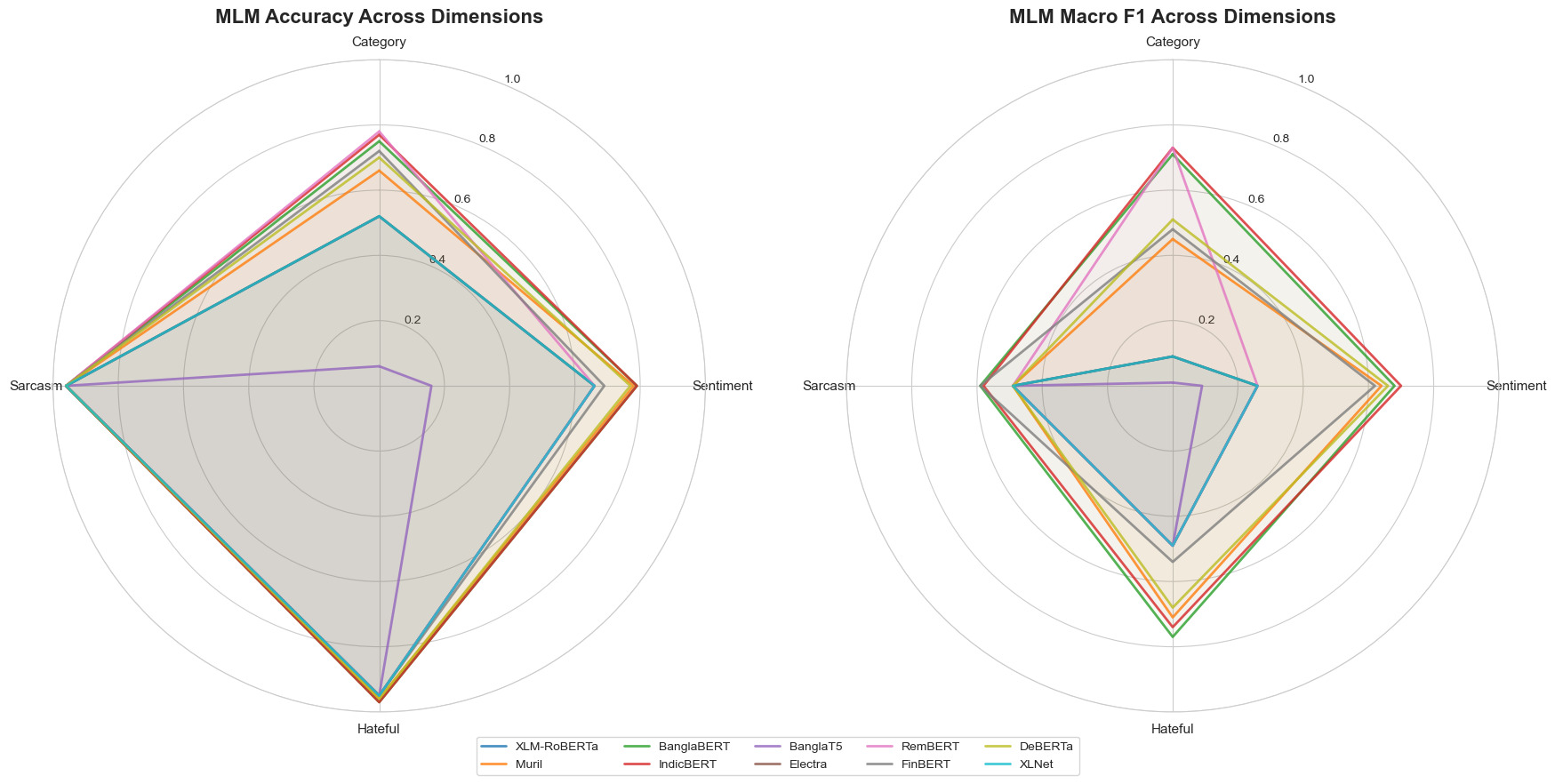}
    \caption{Comparative Radar Charts illustrating Multi/Monolingual Models Sensitivity across Accuracy and Macro F1 dimensions.}
    \label{fig:mlmsen}
\end{figure}
\begin{figure}
    \centering
    \includegraphics[width=\linewidth]{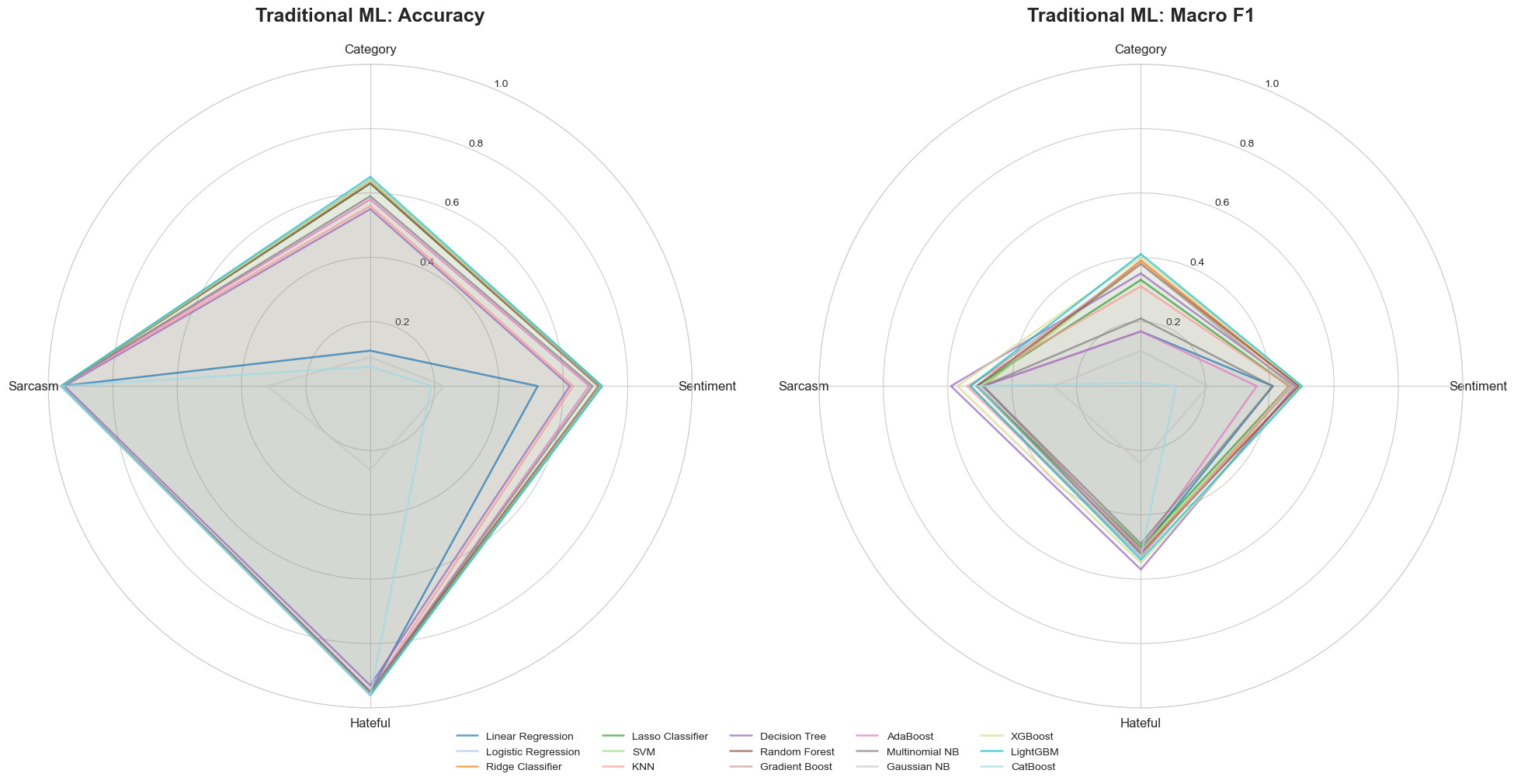}
    \caption{Comparative Radar Charts illustrating Traditional Machine Learning Models Sensitivity across Accuracy and Macro F1 dimensions.}
    \label{fig:mlsen}
\end{figure}

\end{document}